\documentclass[anon,12pt]{clear2023} %

\usepackage{amsmath}
\usepackage{amsthm}
\usepackage[most]{tcolorbox}
\usepackage[ruled,linesnumbered]{algorithm2e}
\usepackage{hhline}
\usepackage{arydshln}
\usepackage{color, colortbl}
\usepackage[labelformat=simple]{subcaption}

\usepackage{soul}

\usepackage{tikz}
\usepackage{pgfplots}
\usepackage{rotating}   
\usetikzlibrary{decorations.pathreplacing,arrows,automata,calc,shapes.geometric}

\DeclareMathOperator*{\argmin}{arg\,min}

\let\oldnl\nl
\newcommand{\nonl}{\renewcommand{\nl}{\let\nl\oldnl}}
\newcommand\ci{\perp\!\!\!\perp}

\makeatletter
\newtheorem*{rep@theorem}{\rep@title}
\newcommand{\newreptheorem}[2]{%
\newenvironment{rep#1}[1]{%
 \def\rep@title{#2 \ref{##1}}%
 \begin{rep@theorem}}%
 {\end{rep@theorem}}}
\makeatother

\pgfmathdeclarefunction{gauss}{2}{%
  \pgfmathparse{1/(#2*sqrt(2*pi))*exp(-((x-#1)^2)/(2*#2^2))}%
}

\definecolor{Gray}{gray}{0.93}

\newtheorem{innercustomgeneric}{\customgenericname}
\providecommand{\customgenericname}{}
\newcommand{\newcustomtheorem}[2]{%
  \newenvironment{#1}[1]
  {%
   \renewcommand\customgenericname{#2}%
   \renewcommand\theinnercustomgeneric{##1}%
   \innercustomgeneric
  }
  {\endinnercustomgeneric}
}

\newcustomtheorem{custom_assumption}{Assumption}

\newtheorem{assumptions1}{Assumption}
\newtheorem{theorem1}{Theorem}

\newreptheorem{theorem}{Theorem}
\newreptheorem{lemma}{Lemma}
\newreptheorem{proposition}{Proposition}

\usepackage{bm,bbm}

\title[Generalizing Clinical Trials with Convex Hulls]{Generalizing Clinical Trials with Convex Hulls}
\usepackage{times}

\clearauthor{%
 \Name{Author Name1} \Email{abc@sample.com}\\
 \addr Address 1
 \AND
 \Name{Author Name2} \Email{xyz@sample.com}\\
 \addr Address 2%
}

\begin{document}

\maketitle

\begin{abstract}
Randomized clinical trials eliminate confounding but impose \textit{strict} exclusion criteria that limit recruitment to a subset of the population. Observational datasets are more inclusive but suffer from confounding -- often providing overly optimistic estimates of treatment response over time due to partially optimized physician prescribing patterns. We therefore assume that the unconfounded treatment response lies somewhere in-between the observational estimate \textit{before} and the observational estimate \textit{after} treatment assignment. This assumption allows us to extrapolate results from exclusive trials to the broader population by analyzing observational and trial data simultaneously using an algorithm called Optimum in Convex Hulls (OCH). OCH represents the treatment effect either in terms of convex hulls of conditional \textit{expectations} or convex hulls (also known as mixtures) of conditional \textit{densities}. The algorithm first learns the component expectations or densities using the observational data and then learns the linear mixing coefficients using trial data in order to approximate the true treatment effect; theory importantly explains \textit{why} this linear combination should hold. OCH estimates the treatment effect in terms both expectations and densities with state of the art accuracy.
\end{abstract}


\section{Introduction}

Randomized clinical trials (RCTs) are the gold standard for inferring causal effects of treatment. RCTs eliminate confounding by randomizing treatment assignment. Randomization however imposes ethical and practical limitations that necessitate \textit{strict} inclusion and exclusion criteria in practice. As a result, trials impose selection bias by limiting entry to a select sub-population. Inferences made with RCTs can thus fail to generalize to everyone seeking help.

Observational datasets, on the other hand, do not randomize treatment assignment. As a result, they suffer from confounding but impose much milder criteria for entry into the study. Inferences made with observational data generalize to the broader population but may not recover the true causal effect no matter how complicated the fit. 

Stated succinctly, observational datasets are inclusive but confounded, whereas RCTs are exclusive but unconfounded. We thus propose to analyze RCT and observational data \textit{simultaneously} in order to eliminate both confounding and selection bias. To this end, we exploit the following observation:
\begin{tcolorbox}[enhanced,frame hidden]
Physicians are like reinforcement learning agents; they identify sub-populations of patients that appear to respond well to a given treatment \textit{over time}, and then prescribe that treatment more often to those patients. The improvement in patient outcomes from before to after treatment assignment is therefore confounded in observational data and usually too large.
\end{tcolorbox}
\noindent Such partially optimized physician prescribing patterns exist before an RCT is ever conducted. \cite{Beach2017} for instance even recommend entire sequences of treatments for different sub-populations based on evidence derived almost exclusively from case reports. 

\begin{figure}[t]
\begin{subfigure}[]{0.5\textwidth}
\begin{center}
\begin{tikzpicture}[mydrawstyle/.style={draw=black, very thick}, x=1mm, y=1mm, z=1mm]

  \draw[thick, ->](0,30)--(0,66);
  \draw[thick, ->](0,30)--(56,30);

  \draw (10,28)--(10,32);
  \draw (40,28)--(40,32);

  \node[] at (10,25) {\smaller Before};
  \node[] at (40,25) {\smaller After};

   \node[rotate=-270] at (-3.5,48){\begin{tabular}{c} \smaller{Outcome}\end{tabular}};

   \node[circle,draw,fill=black,scale=0.5] (a) at (10,40){}; 
   \node[circle,draw,fill=black,scale=0.5] (b) at (40,57){};
   \draw (a) -- (b);

     \draw [decorate,
    decoration = {brace,amplitude=10pt}] (42,57)--(42,40);

    \node[rotate=-90] at (51,48){\begin{tabular}{c} \smaller{RCT estimate}\\ \smaller{in this range} \end{tabular}};

\end{tikzpicture}
\end{center}
\caption{} \label{fig:main_assump:simple}
\end{subfigure}
\begin{subfigure}[]{0.5\textwidth}
\begin{center}
\begin{tikzpicture}[mydrawstyle/.style={draw=black, very thick}, x=1mm, y=1mm, z=1mm]

  \draw[thick, ->](0,30)--(0,66);
  \draw[thick, ->](0,30)--(56,30);

  \draw (10,28)--(10,32);
  \draw (40,28)--(40,32);

  \node[] at (10,25) {\smaller $M=0$};
  \node[] at (40,25) {\smaller $M=1$};

   \node[rotate=0] at (-3.5,48){\begin{tabular}{c} \smaller{$Y$} \end{tabular}};

   \node[circle,draw,fill=black,scale=0.5] (a) at (10,40){}; 
   \node[circle,draw,fill=black,scale=0.5] (b) at (40,57){};
   \draw (a) -- (b);

    \node[] (a1) at (10,36.5){\smaller{$\mathbb{E}^o(Y_0|\bm{X})$}}; 
        \node[] (b1) at (40,60.5){\smaller{$\mathbb{E}^o(Y_1|\bm{X},T)$}}; 

     \draw [decorate,
    decoration = {brace,amplitude=10pt}] (42,57)--(42,40);

    \node[rotate=-90] at (51,48){\begin{tabular}{c} \smaller{$\mathbb{E}^r(Y_1(T)|\bm{X})$}\\ \smaller{in this range} \end{tabular}};

\end{tikzpicture}
\end{center}
\caption{}  \label{fig:main_assump:po}
\end{subfigure}
\caption{Illustration of the main assumption used in this paper. Lines denote the change in response to a given treatment over time in the observational data. (a) The RCT estimate lies in-between the observational estimates seen before and after treatment administration. (b) The same figure as (a) but with potential outcomes notation.}
 \label{fig:main_assump}
\end{figure}
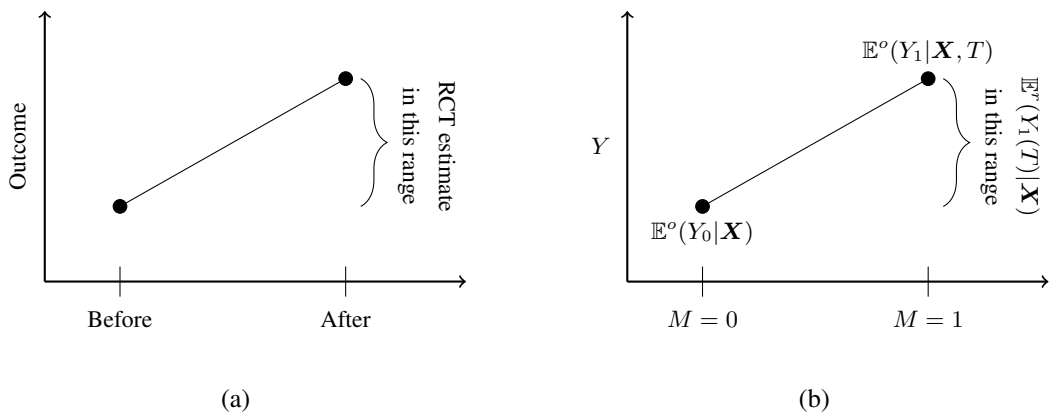

Physicians however also want to isolate the treatment effect from confounding bias using a rigorous RCT. RCTs ignore sub-populations by randomizing treatment assignment, so we expect the RCT estimate of patient outcomes to lie somewhere in-between the observational study estimates seen before and after treatment administration (Figure \ref{fig:main_assump:simple}). For example, observational studies of lithium suggested that the medication decreases suicide attempts by a factor of two to three \citep{Goodwin03,Hayes16}. Physicians often prescribe lithium to chronically suicidal patients in the hopes that the medicine will decrease suicide attempts \textit{in the future}. Large double blinded RCTs of lithium in broader populations replicated the decrease in suicide attempts over time but at a much smaller magnitude \citep{Lauterbach08,Oquendo11,LithiumRCT}. Observational studies therefore yielded an effect size that was in the same direction but too large.

We will convert the above observation into a precise assumption after reviewing the potential outcomes framework. We also develop an algorithm called Optimum in Convex Hulls (OCH) that exploits the highlighted observation by analyzing observational and trial data simultaneously. OCH estimates the treatment effect across the entire population in terms of the difference of two convex hulls. The algorithm can recover the treatment effect in terms of conditional expectations or even conditional densities, so that patients can visualize the relative probabilities of treatment effect across \textit{all} possible outcome values. OCH yields state of the art accuracy in practice.

\section{Potential Outcomes}

We assume binary treatment assignment denoted by the random variable $T$ with $T=0$ or $T=1$, and $\mathbb{P}(T)>0$. We also adopt the potential outcomes framework, where we assume the existence of two potential outcomes $Y(0)$ and $Y(1)$ for all patients. However, we can only observe the single outcome $Y(t)$ for the subject assigned to $T=t$.

Let $\bm{X}$ denote the set of patient covariates measured prior to treatment assignment. Randomizing treatment assignment over the \textit{entire} population ensures that we have $\{Y(0),Y(1)\} \ci T | \bm{X}$ and\\ $\bm{X} \ci T$, so that treatments are given regardless of potential outcomes and patient characteristics. However, patients and clinicians may find randomization disturbing because they cannot ensure optimal treatment assignment. RCTs therefore impose \textit{strict} exclusion criteria\footnote{RCTs also utilize inclusion criteria, which we can convert to exclusionary ones by logical negation.} in practice based on pre-treatment covariates $\bm{X}$ that limit entry into the study. For example, an RCT examining the effects of anti-depressants may exclude patients with severe alcohol use, since these patients are more likely to benefit from inpatient detoxification. Without loss of generality, let $S$ denote a binary random variable representing selection bias (not a study indicator); the variable takes on a value of one if a patient is included in the RCT and zero otherwise.

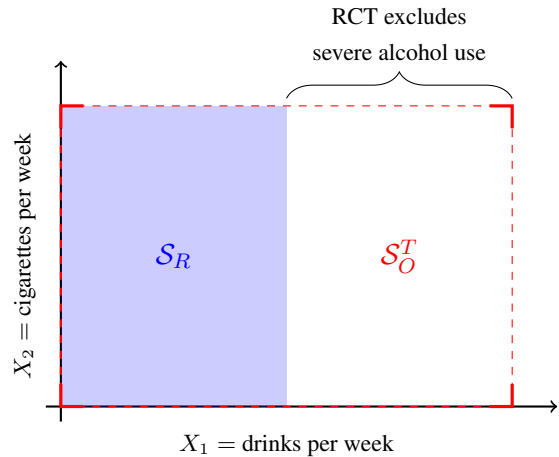
\begin{figure}[t]
\begin{center}
\begin{tikzpicture}[mydrawstyle/.style={draw=black, very thick}, x=1mm, y=1mm, z=1mm]

  \draw[thick, ->](0,28)--(0,76);
  \draw[thick, ->](-2,30)--(66,30);
  
  \fill[fill=blue, opacity=0.2](0,30)--(0,70)--(30,70)--(30,30)--(0,30);

  \draw[mydrawstyle, draw=red](0,33)--(0,30)--(3,30);
  \draw[mydrawstyle, draw=red](3,70)--(0,70)--(0,67);
  \draw[mydrawstyle, draw=red](57,30)--(60,30)--(60,33);
  \draw[mydrawstyle, draw=red](57,70)--(60,70)--(60,67);
  
  \draw[draw=red, thick, dashed](0,30)--(0,70);
  \draw[draw=red, dashed](60,30)--(60,70);
  \draw[draw=red, dashed](0,70)--(60,70);
  \draw[draw=red, dashed](0,30)--(60,30);
  
  \node[] at (15,50){\textcolor{blue}{$\mathcal{S}_R$}};
  \node[] at (45,50){\textcolor{red}{$\mathcal{S}_O^T$}};
  
  \node[] at (30,25){\begin{tabular}{c} \smaller{$X_1=$ drinks per week} \end{tabular}};
  
  \node[rotate=-270] at (-5,50){\begin{tabular}{c} \smaller{$X_2=$ cigarettes per week} \end{tabular}};
  
  \draw [decorate,
    decoration = {brace,amplitude=10pt}] (30,71)--(60,71) node at (45,79.5) {\begin{tabular}{c} \smaller{RCT excludes}\\ \smaller{severe alcohol use} \end{tabular}};
  
\end{tikzpicture}
\end{center}
\caption{Example of an RCT strictly excluding patients who drink alcohol excessively. The support $\mathcal{S}_O^T$ is outlined with a dashed red line and $\mathcal{S}_R$ is shaded in blue; notice that we have $\mathcal{S}_R \subset \mathcal{S}_O^T$.}  \label{fig:alcohol}
\end{figure}

RCTs impose selection bias, but they still randomize treatment assignment \textit{among the recruited} so that we have:
\begin{assumptions1} \label{assump:randomization}
$\{Y(0),Y(1)\} \ci T | \{\bm{X},S=1\}$ and $\bm{X} \ci T | S=1$ in the RCT distribution.
\end{assumptions1}
\noindent Trials therefore eliminate confounding but only for the recruited sub-group. These independence relations do not hold in the observational distribution in general. RCTs then sample patient covariates from the distribution $\mathbb{P}(\bm{X}|T,S=1)=\mathbb{P}(\bm{X}|S=1)$ with support $\mathcal{S}_R$, while observational studies sample from the distribution $\mathbb{P}(\bm{X}|T)=\mathbb{E}_{S|T}[\mathbb{P}(\bm{X}|T,S)]$ with support $\mathcal{S}^T_O$. We also have:
\begin{assumptions1} \label{assump:support}
$\mathcal{S}_R \subseteq \mathcal{S}^T_O$,
\end{assumptions1}
\noindent since RCTs impose selection bias with exclusion criteria whereas observational datasets do not. We provide an illustration in Figure \ref{fig:alcohol} using the anti-depressant example, where patients with severe alcohol use are excluded from the RCT. We would like to draw conclusions about this sub-population as well because some of these patients drink large amounts of alcohol to cope with depression. Any procedure that makes inferences on $\mathcal{S}_R$ must therefore \textit{extrapolate} to $\mathcal{S}^T_O \setminus \mathcal{S}_R$ in order to generalize to the broader population.

Since RCTs construct their exclusion criteria based on $\bm{X}$, we also frequently have:
\begin{assumptions1}  \label{assump:ignoreS}
$\{Y(0),Y(1)\} \ci S | \bm{X}$ in the RCT distribution.
\end{assumptions1}
\noindent For instance, $\bm{X}$ may include amount of alcohol use in the aforementioned example, and patients who exceed a certain threshold are excluded from the study. $S$ therefore provides no additional information on a potential outcome given $\bm{X}$: $\mathbb{P}(Y(T)|\bm{X},S) = \mathbb{P}(Y(T)|\bm{X})$. Note that this assumption only applies to $\mathcal{S}_R$. We provide an overview of Assumptions \ref{assump:randomization}-\ref{assump:ignoreS} in terms of graphical models in the Appendix \ref{app:graph}.

In summary, RCTs impose selection bias (Assumption \ref{assump:support}), often using predetermined criteria (Assumption \ref{assump:ignoreS}), but eliminate confounding among the recruited (Assumption \ref{assump:randomization}). In contrast, observational studies eliminate selection bias (Assumption \ref{assump:support}) but introduce confounding. We therefore focus on eliminating both selection bias and confounding by analyzing RCT and observational data simultaneously. We in particular aim to estimate the \textit{conditional average treatment effect} (CATE) given by $g(\bm{X}) \triangleq \mathbb{E}(Y(1)|\bm{X}) - \mathbb{E}(Y(0)|\bm{X})$ for everyone, or on $\mathcal{S}_O=\mathcal{S}_O^1 \cap \mathcal{S}_O^0$. The CATE corresponds to a difference of two conditional expectations, where we condition on $\bm{X}$ in order to identify patient-specific treatment effects in the spirit of precision medicine. 

\begin{figure}[h]
\begin{center}
\begin{tikzpicture}[mydrawstyle/.style={draw=black, very thick}, x=1mm, y=1mm, z=1mm,scale=0.8]

\begin{axis}[
  no markers, domain=-2.5:3.5, ymin=0,ymax=0.7, samples=100,
  axis x line*=middle, axis y line*= left, xlabel=$y$,
  every axis x label/.style={at={(axis description cs:0.95,-0.1)}},
  height=5cm, width=12cm, xtick = \empty, ytick=\empty,
  enlargelimits=false, clip=false, axis on top,
  grid = none
  ]
  
  \addplot [very thick,blue,opacity=0.6] {gauss(1.5,0.5)};
  
  \addplot [very thick,blue,opacity=0.6] {gauss(-0.1,0.7)};
  
  \node[circle,red,fill,inner sep=1.5pt] at (axis cs:0.7,0) {};

  \end{axis}
  
  \node[circle,red,fill,inner sep=1.5pt] at (7,26) {};
 
  \node[] at (15,26) {\smaller CATE};
  
  \fill[blue,opacity=0.6](6,19.7)--(6,20.2)--(8,20.2)--(8,19.7)--(6,19.7);

    \node[] at (15,20) {\smaller CDTE};

\end{tikzpicture}
\end{center}
\caption{The CATE only provides a point estimate for each patient, whereas the CDTE summarizes the probabilities across \textit{all} possible outcome values. We do not require Gaussianity for the CDTE.}  \label{fig:CATEvCDTE}
\end{figure}
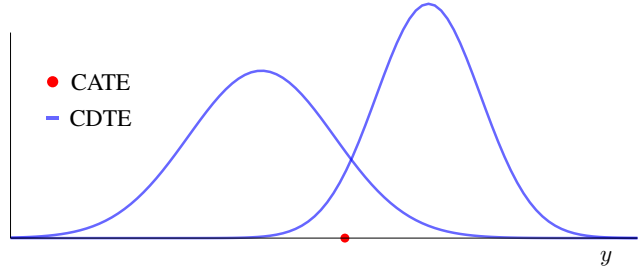

The CATE unfortunately only provides a point estimate for each patient and therefore does not take into account the uncertainty in the outcome value. Patients understand this uncertainty and frequently want to know the probabilities associated with all possible outcomes. We therefore also seek to recover \textit{conditional densities of treatment effect} (CDTE), or $p(Y(1)|\bm{X})$ and $p(Y(0)|\bm{X})$ on $\mathcal{S}_O$. The CDTE summarizes the probabilities associated with all possible outcome values, i.e. $p(Y(1)=y|\bm{X})$ and $p(Y(0)=y|\bm{X})$ for all possible $y$ (Figure \ref{fig:CATEvCDTE} in Appedix \ref{app:CATEvCDTE}). We will recover the CATE and CDTE even for patients excluded from the trial by analyzing RCT and observational data simultaneously. 

\section{Main Assumptions}\label{sec:main_assump}

We now introduce the main assumptions used in this paper. Unlike Assumptions \ref{assump:randomization}-\ref{assump:ignoreS}, the assumptions proposed in this section impose functional restrictions that may not map onto changes in graphical structure. We will consider two time steps: before and after treatment assignment, corresponding to the binary random variable $M=0$ and $M=1$, respectively. We use the notation $Y_M(T)$ to denote the potential outcome with treatment assignment $T$ at time step $M$. Treatment cannot causally affect the outcome before treatment assignment, or at $M=0$. 

\subsection{Conditional Expectations}

Let $Y_M = TY_M(1) + (1-T)Y_M(0)$ denote the observed potential outcome. We can then write the conditional expectation in the observational distribution as  $\mathbb{E}^o(Y_M(T)|\bm{X},T)=\mathbb{E}^o(Y_M|\bm{X},T)$ and that in the ideal RCT distribution, where everyone is randomized, as $\mathbb{E}^r(Y_M(T)|\bm{X})$ because $Y_M(T) \ci T |\bm{X}$ in this case. The superscripts emphasize the observational and RCT distributions.

We in general have $\mathbb{E}^o(Y_0|\bm{X},T=0) \not =\mathbb{E}^o(Y_0|\bm{X},T=1)$ in observational studies even before treatment assignment due to confounding. Physicians may for instance choose to give a stronger medication $T=1$ to sicker patients. Sicker patients have worse outcomes even before treatment assignment as reflected by the random variable $Y_0(1)$. However, the quantity:
\begin{equation} \label{eq:no_effect}
    \mathbb{E}^o(Y_0|\bm{X}) = \sum_{t\in \{0,1\}}\mathbb{E}^o(Y_0|\bm{X},T=t)\mathbb{P}^o(T=t|\bm{X})
\end{equation}
provides a baseline value for \textit{all} patients with covariates $\bm{X}$ before treatment has time to take effect, or regardless of whether $T=1$ or $T=0$ at time step $M=0$.

As alluded to in the introduction, observational studies tend to produce overly large estimates of treatment response over time, so the treatment response in the RCT should lie somewhere in-between the observational estimate before and after treatment assignment. We now more explicitly assume:
\begin{assumptions1} \label{assump:main_e}
$\mathbb{E}^r(Y_1(T)|\bm{X}) = \mathbb{E}^o(Y_1|\bm{X},T)\mu_T + \mathbb{E}^o(Y_0|\bm{X}) (1-\mu_T)$ where $\mu_T \in [0,1]$.
\end{assumptions1}
\noindent The above equality holds for $T=0$ and $T=1$, and we may have $\mu_0 \not = \mu_1$. In other words, $\mathbb{E}^r(Y_1(T)|\bm{X})$ in the RCT distribution lies in the convex hull of $\mathbb{E}^o(Y_0|\bm{X})$ and $\mathbb{E}^o(Y_1|\bm{X},T)$ for before and after treatment assignment, respectively (Figure \ref{fig:main_assump:po}). 

Physicians can detect effective treatments in sub-populations by observing the change in their patients before and after treatment administration. We therefore do not expect treatment in the RCT to have an opposite effect over time, or obtain $\mu_T < 0$. We similarly do not expect randomization to outperform the intelligent selection of treatment by trained professionals, or obtain $\mu_T > 1$. However, clinicians then start prescribing certain medications to select patients based on baseline characteristics $\bm{X}$, so confounding may account for most of the improvement in the mean outcome seen in observational data -- implying that $\mu_T \leq 1$. Assumption \ref{assump:main_e} formalizes these ideas within the potential outcomes framework. We will also see that enforcing $\mu_T \in [0,1]$, as opposed to allowing $\mu_T \in \mathbb{R}$, increases the robustness of the proposed algorithm to stricter exclusion criteria in Section \ref{sec:synth}.

 We do \textit{not} simply assume $g(\bm{X}) = \gamma[\mathbb{E}^o(Y_1|\bm{X},T=1)-\mathbb{E}^o(Y_1|\bm{X},T=0)]$ with $\gamma \in [0,1]$, so that the CATE either matches or underperforms the observational estimate at $M=1$. Assumption \ref{assump:main_e} uses the \textcolor{blue}{magnitude of change} within each treatment in congruence with how physicians choose a treatment that optimizes patient outcome \textcolor{blue}{over time} like reinforcement learning agents. The CATE can have a larger magnitude than the observational estimate at $M=1$ under Assumption \ref{assump:main_e}; see for example Figure \ref{fig:ex:bigger} in Appendix \ref{app:CE}.

  Assumption \ref{assump:main_e} similarly differs from the Manski bounds \citep{Manski89}, where $\mathbb{E}^r(Y_1(T)|\bm{X})$ is bounded above by $\mathbb{E}^o(Y_1|\bm{X},T) \mathbb{P}^o(T|\bm{X}) + \mathbb{P}^o(\neg T|\bm{X})$ and below by $\mathbb{E}^o(Y_1|\bm{X},T) \mathbb{P}^o(T|\bm{X})$ when $Y_1(T) \in [0,1]$. Assumption \ref{assump:main_e} again bounds changes \textit{over time}, whereas the Manski bounds only apply to $M=1$.

 Finally, Assumption \ref{assump:main_e} differs from the parallel slopes assumption used in conditional Difference in Differences \citep{Card93,Abadie05}. The parallel slopes assumption enforces the equality $\mathbb{E}^o(Y_1(0)|\bm{X},T=1) - \mathbb{E}^o(Y_0(0)|\bm{X},T=1) =  \mathbb{E}^o(Y_1(0)|\bm{X},T=0) - \mathbb{E}^o(Y_0(0)|\bm{X},T=0)$, or \textit{equalizes} the slopes of change across treatment assignments. Difference in Differences then identifies the conditional average treatment effect \textit{on the treated}. In contrast, Assumption \ref{assump:main_e} bounds the RCT estimate using the slopes of change in the observational distribution. We will exploit Assumption \ref{assump:main_e} in Section \ref{sec:OCH} to identify the CATE irrespective of treatment assignment.

\subsection{Conditional Densities}

We can extend Assumption \ref{assump:main_e} to conditional densities as follows:
\begin{assumptions1} \label{assump:main_p}
$p^r(Y_1(T)|\bm{X}) = \underbrace{p^o(Y_1|\bm{X},T)}_{(1)}\mu_T + \underbrace{p^o(Y_0|\bm{X})}_{(2)} (1-\mu_T)$ where $\mu_T \in [0,1]$,
\end{assumptions1}
\noindent so that we no longer restrict ourselves to a convex hull of the expectations but to a mixture of two conditional densities. Clearly Assumption \ref{assump:main_p} implies Assumption \ref{assump:main_e} but not vice versa. Assumption \ref{assump:main_p} also implies that we can decompose the density $p^r(Y_1(T)|\bm{X})$ into two groups of patients: (1) those who respond to treatment just like in the observational dataset, and (2) those who do not respond. The coefficients $\mu_T$ and $1-\mu_T$ represent learnable parameters that correspond to the unknown proportion of patients who satisfy (1) and (2), respectively.

\section{Optimum in Convex  Hulls}\label{sec:OCH}

We want to isolate the effect of treatment from the confounding bias introduced by partially optimized physician prescribing patterns. Unfortunately, Assumptions \ref{assump:main_e} and \ref{assump:main_p} only bound the treatment response over time between two conditional expectations or densities, respectively. We thus present algorithms that pinpoint the \textit{exact} values for the CATE. See Appendix \ref{app:CDTE} for the derivation of the OCH$_d$ algorithm that pinpoints the CDTE.

\subsection{CATE with Two Time Steps}

We first consider the ideal scenario, where we have access to two time steps worth of observational data. The CATE is equivalent to the following under Assumption \ref{assump:main_e}:
\begin{equation*}
\begin{aligned}
    &\hspace{5mm}\mathbb{E}^r(Y_1(1)|\bm{X}) - \mathbb{E}^r(Y_1(0)|\bm{X})\\
    &\stackrel{4}{=} \Big[\mathbb{E}^o(Y_1|\bm{X},T=1) \mu_1 + \mathbb{E}^o(Y_0|\bm{X})(1-\mu_1)\Big] - \Big[\mathbb{E}^o(Y_1|\bm{X},T=0) \mu_0 + \mathbb{E}^o(Y_0|\bm{X}) (1-\mu_0)\Big]\\
    &= \psi_{\mu}(\bm{X}) 
\end{aligned}
\end{equation*}
\noindent for $\mu = (\mu_0,\mu_1) \in [0,1]^2$. The number above the equality sign references the assumption. Note that $\psi_{\mu}(\bm{X})$ applies to all of $\mathcal{S}_O$, since the three conditional expectations on the right hand side are derived from the observational distribution.

The quantities $\mu_0$ and $\mu_1$ however remain unknown. Fortunately, the CATE is equivalent to $\mathbb{E}^r(Y_1(1)|\bm{X},T=1,S=1) - \mathbb{E}^r(Y_1(0)|\bm{X},T=0,S=1)$ in the RCT distribution under Assumptions \ref{assump:randomization} and \ref{assump:ignoreS} because $\mathbb{E}^r(Y_1(1)|\bm{X},T=1,S=1) - \mathbb{E}^r(Y_1(0)|\bm{X},T=0,S=1)
    \stackrel{1}{=} \mathbb{E}^r(Y_1(1)|\bm{X},S=1) - \mathbb{E}^r(Y_1(0)|\bm{X},S=1)
    \stackrel{3}{=} \hspace{1mm} \mathbb{E}^r(Y_1(1)|\bm{X}) - \mathbb{E}^r(Y_1(0)|\bm{X})$. We can therefore fit $\mu_0$ and $\mu_1$ on the area of overlap $\mathcal{S}_O \cap \mathcal{S}_R = \mathcal{S}_R$ by Assumption \ref{assump:support} using the trial data. We in particular minimize the distance to the CATE on $\mathcal{S}_R$ by solving:
\begin{equation} \label{eq:optCATE}
\begin{aligned}
    \mu^* = \argmin_{\mu} &\hspace{1mm} \mathbb{E}_{\bm{X}|S=1}^r\Big( g(\bm{X}) - \psi_{\mu}(\bm{X}) \Big)^2\\
    & s.t. \hspace{2mm} 0 \leq \mu \leq 1,
\end{aligned}
\end{equation}
where the outer expectation is taken over $\mathcal{S}_R$. We are now ready to state a main result:
\begin{theorem1} \label{thm:CATE}
$\psi_{\mu^*}(\bm{X})$ is equivalent to the CATE on $\mathcal{S}_O$ under Assumptions \ref{assump:randomization}-\ref{assump:main_e}.
\end{theorem1}
\begin{proof}
The CATE is equivalent to $\psi_{\mu}(\bm{X})$ on $\mathcal{S}_O$ for some $\mu \in [0,1]^2$ by Assumption \ref{assump:main_e}. The CATE is also equivalent to $\mathbb{E}^r(Y_1(1)|\bm{X},T=1,S=1) - \mathbb{E}^r(Y_1(0)|\bm{X},T=0,S=1)$ in the RCT distribution. The quantity $\psi_{\mu}(\bm{X})$ is unique on $\mathcal{S}_O$ for any $\mu \in \mathbb{R}^2$. The solution $\mu^*$ solving Expression \eqref{eq:optCATE} is unique on $\mathcal{S}_R$ because $\mathcal{S}_O \cap \mathcal{S}_R = \mathcal{S}_R$ by Assumption \ref{assump:support}. The quantity $\psi_{\mu^*}(\bm{X})$ is therefore equivalent to the CATE on $\mathcal{S}_O$.
\end{proof}

We of course must estimate all necessary conditional expectations, in addition to $\mu_0$ and $\mu_1$, using the observational and trial data. Estimating the necessary conditional expectations and $\mu$ leads to the OCH$_2$ algorithm summarized in Algorithm \ref{alg:OCH2}. OCH$_2$ first estimates the CATE on $\mathcal{S}_R$ using the RCT data in Step \ref{alg:OCH2:RCT}. The algorithm then estimates each entry of $H(\bm{X}) = \{ \mathbb{E}^o(Y_1|\bm{X},T=1), \mathbb{E}^o(Y_1|\bm{X},T=0),\mathbb{E}^o(Y_0|\bm{X}) \}$ using the observational data in Step \ref{alg:OCH2:H}. Next, OCH$_2$ approximates $\mu^*$ using the trial data in Step \ref{alg:OCH2:mu} by solving the empirical version of Expression \eqref{eq:optCATE} with the solutions of Steps \ref{alg:OCH2:RCT} and \ref{alg:OCH2:H}:
\begin{equation} \label{eq:optCATE_emp}
\begin{aligned}
    \widehat{\mu} = \argmin_{\mu} &\hspace{1mm} \frac{1}{2n} \sum_{i=1}^{2n} \Big( \overbrace{\widehat{g}(\bm{x}_i)}^{\textnormal{Step 1}} - \overbrace{\widehat{\psi}_{\mu}(\bm{x}_i)}^{\textnormal{Step 2}} \Big)^2\\
    &\hspace{10mm} s.t. \hspace{2mm} 0 \leq \mu \leq 1,
\end{aligned}
\end{equation}
where $\widehat{\psi}_{\mu}(\bm{x}_i) = \Big[\widehat{\mathbb{E}}^o(Y_1|\bm{x}_i,T=1) \mu_1 + \widehat{\mathbb{E}}^o(Y_0|\bm{x}_i)(1-\mu_1)\Big] - \Big[\widehat{\mathbb{E}}^o(Y_1|\bm{x}_i,T=0) \mu_0 + \widehat{\mathbb{E}}^o(Y_0|\bm{x}_i) (1-\mu_0)\Big]$, and $n$ refers to the RCT sample size per treatment -- assumed to be the same per treatment for notational convenience. The algorithm finally predicts $\widehat{\psi}_{\widehat{\mu}}(\bm{x})$ for all $\bm{x}$ in the test set $\mathcal{T}$ each lying anywhere in $\mathcal{S}_O$.

\begin{algorithm}[]
 \nonl \textbf{Input:} trial data, observational data, test points $\mathcal{T}$\\
 \nonl \textbf{Output:} $\widehat{\psi}_{\widehat{\mu}}(\bm{X})$ on $\mathcal{T}$\\
 \BlankLine

Estimate $g(\bm{X})$ on $\mathcal{S}_R$ using the trial data \label{alg:OCH2:RCT}\\
Estimate each entry of $H(\bm{X})$ on $\mathcal{S}_O$ using the observational data \label{alg:OCH2:H}\\
Solve Expression \eqref{eq:optCATE_emp} using $\widehat{g}(\bm{X})$ and $\widehat{H}(\bm{X})$ on the trial data \label{alg:OCH2:mu}\\
Predict $\widehat{\psi}_{\widehat{\mu}}(\bm{X})$ on $\mathcal{T}$ \label{alg:OCH2:test}

 \caption{Optimum in Convex Hulls with Two Time Steps (OCH$_2$)} \label{alg:OCH2}
\end{algorithm}

\subsection{CATE with One Time Step}

We unfortunately do not always have access to two time steps of observational data. Suppose however that the potential outcomes $Y_M(T)$ for $M=0,1$ and $T=0,1$ are appropriately normalized so that they are bounded below by zero; we can almost always satisfy this condition in clinical practice. We then have $\mathbb{E}(Y_0|\bm{X}) \geq 0$. If we take the worst case scenario $\mathbb{E}(Y_0|\bm{X}) = 0$, then Assumption \ref{assump:main_e} boils down to:
\begin{custom_assumption}{4$^\prime$} \label{assump:main_e2}
$\mathbb{E}^r(Y_1(T)|\bm{X}) = \mathbb{E}^o(Y_1|\bm{X},T)\mu_T$ where $\mu_T \in [0,1]$.
\end{custom_assumption}
\noindent In other words, the above statement relaxes Assumption \ref{assump:main_e} from the convex hull of $\mathbb{E}^o(Y_1|\bm{X},T)$ and $\mathbb{E}^o(Y_0|\bm{X})\geq 0$ to the wider convex hull of $\mathbb{E}^o(Y_1|\bm{X},T)$ and $0$. Recovering the CATE then proceeds exactly as in Algorithm \ref{alg:OCH2}, but by setting $\widehat{\mathbb{E}}^o(Y_0|\bm{X})$ to zero. We refer to this variant as OCH$_1$ for one time step.

\section{Experiments}

We now investigate the accuracy of OCH using both synthetic and real data. Code is available at https://github.com/ericstrobl/OCH.

\subsection{Algorithms}

\textit{State of the Art.} See Appendix \ref{app:RW} for a comprehensive discussion of related work. We compare OCH$_2$ against (1a) OCH$_1$ as well as six other algorithms representing the state of the art in CATE estimation under strict exclusion criteria: (2a) regression with RCT data only, (3a) regression with observational data only, (4a) OLT \citep{Jackson17}, (5a) 2Step \citep{Kallus18}, (6a) the conditional version of Difference in Differences (CDD) \citep{Abadie05}, (7a) SDD \citep{Strobl21}. We compare OCH$_d$ for CDTE estimation with (1b) conditional density estimation with RCT data only and (2b) conditional density estimation with observational data only, since all other algorithms can only estimate the CATE. We will use the acronym RCT to refer to (2a) or (1b), and OBS to refer to (3a) or (2b), when it is clear that we mean the algorithms and not the datasets.\\

\noindent \textit{Ablation Studies.} 
OCH$_1$ is an ablated version of OCH$_2$ obtained by removing the pre-treatment time step. We also compare the OCH variants for the CATE against (8a) UNC$_2$, or OCH$_2$ with the constraint in Expression \eqref{eq:optCATE_emp} removed, and similarly (9a) UNC$_1$, or OCH$_1$ with the constraint removed. For the CDTE, we introduce (3b) UNC$_d$, or OCH$_d$ with the constraint in Expression \eqref{eq:optCDTE_emp} removed.\\

Note that the algorithms use different machine learning algorithms to estimate the required conditional expectations or densities out of box. We are however interested in isolating the performance of each algorithm independent of the chosen regressor or conditional density estimator. We therefore instantiate all algorithms with kernel ridge regression to estimate the required conditional expectations and the least squares probabilistic classifier (discrete outcome) or Dirac delta regression (continuous outcome) to estimate the required conditional densities in non-parametric form \citep{Yamada11,Strobl21ddr}. We equip both methods with the infinite knot spline kernel \citep{Vapnik98,Izmailov13}. We select the $\lambda$ hyperparameter for kernel ridge regression from the set $\{$1E-8, 1E-7, $\dots$, 1E-1$\}$ and otherwise use default hyperparameters for the least squares probabilistic classifier and Dirac delta regression.

\subsection{Synthetic Data} \label{sec:synth}

\subsubsection{Simulation} 

We generate synthetic data using a mixture model. We sample the observational data i.i.d. from the following distribution:
\begin{equation*}
\begin{aligned}
    &Y_M(T) \sim \mathcal{N}(f_{MT}(Z),0.1)
\end{aligned}
\end{equation*}
with $Z = \sum_{i=1}^p X_i$, each $X_i \sim \mathcal{U}(-1,1)$ and the function $f_{MT}(Z)$ sampled uniformly from the set $\{Z,Z\Psi(Z),\textnormal{exp}(-Z^2),\textnormal{tanh}(Z) \}$ for $M=0,1$ and $T=0,1$. We sample the RCT data from:
\begin{equation*}
\begin{aligned}
    &Y_1(T) \sim \mu_T \mathcal{N}(f_{1T}(Z),0.1) + (1-\mu_T)\mathcal{N}_2(Z),
\end{aligned}
\end{equation*}
where $\mu_T \sim \mathcal{U}(0,1)$ and $\mathcal{N}_2(Z) = \frac{1}{2}\mathcal{N}(f_{01}(Z),0.1) + \frac{1}{2} \mathcal{N}(f_{00}(Z),0.1)$. The density $p^r(Y_1(T)|\bm{X})$ is therefore a mixture of $p^o(Y_1|\bm{X},T)$ and $p^o(Y_0|\bm{X}) = \frac{1}{2}p^o(Y_0|\bm{X},T=1) + \frac{1}{2}p^o(Y_0|\bm{X},T=0)$ satisfying Assumptions \ref{assump:main_e} and \ref{assump:main_p}. See Appendix \ref{app:sim:viol} for simulation results when the assumptions are violated.

We generate 1000 samples for the observational data split evenly between the two treatments and two time steps. We also generate 100 samples for the trial data split evenly between the two treatments. We impose strict inclusion criteria onto the trial data by excluding $r = 0, 25, 50, 75, 90$ or $95\%$ of patients by sampling $X_1$ according to $\mathcal{U}(-1+0.02r,1)$; for example, excluding $50\%$ of patients is equivalent to sampling from $\mathcal{U}(0,1)$ on $50\%$ of the support of $\mathcal{U}(-1,1)$. We repeat the above procedure $500$ times for the CATE with the excluded percentages and $p=1,2,6$ or $10$ variables in $\bm{X}$. We therefore generate a total of $500 \times 6 \times 4 = 12000$ independent datasets. We also repeat the above procedure $100$ times for the CDTE for a total of $100 \times 6 \times 4 = 2400$ datasets. We finally compare the algorithms by either computing the median of the mean squared error (MSE) to the ground truth CATE, or the median of the MISE to the ground truth CDTE; we use the median instead of the mean because the MSE and MISE histograms are skewed to the right for some algorithms.

\subsubsection{Performance} 

\begin{table*}[t]
\begin{center}
\begin{subtable}{0.98\textwidth}  
\centering
\begin{tabular}{ccccccc}
\hhline{=======}
      & 0\%             & 25              & 50              & 75              & 90              & 95              \\ \hline
\rowcolor{Gray} OCH$_2$   & \textbf{0.0449} & \textbf{0.0503} & \textbf{0.0529} & \textbf{0.0550} & \textbf{0.0550} & \textbf{0.0609} \\
\rowcolor{Gray} OCH$_1$  & 0.0555          & \textbf{0.0564}          & \textbf{0.0564}         & \textbf{0.0589}          & \textbf{0.0578}          & \textbf{0.0649}          \\
\rowcolor{Gray} UNC$_2$   & \textbf{0.0520} & 0.0653          & 0.0710          & 0.0819          & 0.0996          & 0.1225\\
\rowcolor{Gray} UNC$_1$   & 0.0606 & 0.0664          & 0.0695          & 0.0827          & 0.0962          & 0.1067\\
SDD   & 0.1266          & 0.1266          & 0.1364         & 0.1453          & 0.1569          & 0.1699          \\
2Step & 0.2146          & 0.2136          & 0.2333         & 0.2634          & 0.2848          & 0.2879          \\
OBS   & 0.2560          & 0.2492          & 0.2491          & 0.2570          & 0.2591          & 0.2631         \\
RCT   & 0.1507          & 0.1949          & 0.2531          & 0.3331          & 0.3724          & 0.3615          \\
OLT   & 0.2606          & 0.3226          & 0.4232          & 0.5204          & 0.5829          & 0.6799          \\
CDD   & 0.6399          & 0.6476          & 0.6145          & 0.6267          & 0.6201         & 0.6514          \\ 
\hhline{=======}
\end{tabular}\caption{}\label{exp:CATE:prop}
\end{subtable}
\end{center}

\vspace{3mm}\begin{subtable}{0.98\textwidth}  
\centering
\captionsetup{justification=centering,margin=2cm}
\begin{tabular}{ccccccc}
\hhline{=======}
      & 0\%             & 25              & 50              & 75              & 90              & 95              \\ \hline
\rowcolor{Gray} OCH$_d$   & \textbf{0.1440} & \textbf{0.1376} & \textbf{0.1417} & \textbf{0.1401} & \textbf{0.1374} & \textbf{0.1461} \\
\rowcolor{Gray} UNC$_d$   & \textbf{0.1448} &  \textbf{0.1384}          & \textbf{0.1442}          & \textbf{0.1426}          & \textbf{0.1410}          &  \textbf{0.1477}\\
OBS   & 0.2304          & 0.2354          & 0.2482          & 0.2434          & 0.2340         & 0.2392          \\
RCT   & 0.2566          & 0.2718          & 0.3128          & 0.3839          & 0.4529          & 0.4560          \\
\hhline{=======}
\end{tabular}
\caption{} \label{table:CDTE:prop}
\end{subtable}
\caption{Accuracy results for (a) the CATE and (b) the CDTE approximately sorted from best to worst. Lower is better. The OCH variants outperform the state of the art across all percentages of excluded subjects. Ablation studies reveal that the regression constraints $\mu_T \in [0,1]$ but not the pre-treatment data are necessary to achieve optimal performance with increasing exclusivity.} \label{exp:CATE_CDTE:prop}
\end{table*}

\textit{Accuracy.} We summarize the results for the CATE in Table \ref{exp:CATE:prop} with algorithms roughly sorted from best to worst. Bolded values correspond to the best performance according to Mood's median test at a Bonferronni corrected $p$-value threshold of 0.05/9, since we ultimately compare each OCH variant against 9 other algorithms. When Assumption \ref{assump:main_e} holds, both OCH$_2$ and OCH$_1$ outperform all of their predecessors (2a-7a) across all percentages of excluded subjects (Table \ref{exp:CATE:prop}) and all numbers of variables in $\bm{X}$ (Table \ref{exp:CATE:vars} in Appendix \ref{app:sim:add}). The algorithms even outperform RCT only with zero percent excluded subjects by taking advantage of the larger sample size of the observational dataset. Moreover, the constraints in OCH$_2$ and OCH$_1$ improve performance in most cases. Similarly, OCH$_2$ and OCH$_1$ outperform all other algorithms except their unconstrained variants when Assumption \ref{assump:main_e} is violated (Tables \ref{exp:CATE:propv} and \ref{exp:CATE:varsv} in Appendix \ref{app:sim:viol}).  

We summarize results for the CDTE in Table \ref{table:CDTE:prop}. OCH$_d$ outperforms its competitors by a large margin regardless of whether Assumption \ref{assump:main_p} holds. The constraints however add little value when estimating the CDTE; OCH$_d$ and UNC$_d$ perform comparably across all proportions of excluded patients (Table \ref{table:CDTE:prop}, and Table \ref{table:CDTE:propv} in Appendix \ref{app:sim:viol}) and across most variable numbers (Table \ref{table:CDTE:vars} in Appendix \ref{app:sim:add} and Table \ref{table:CDTE:varsv} in Appendix \ref{app:sim:viol}). Densities must be non-negative and integrate to one, so constraining the mixing coefficients offers some but ultimately minimal additional benefit.\\

\begin{figure}[h]
\begin{subfigure}[]{0.5\textwidth}
\captionsetup{justification=centering,margin=2cm}
    \centering
    \includegraphics[scale=0.65]{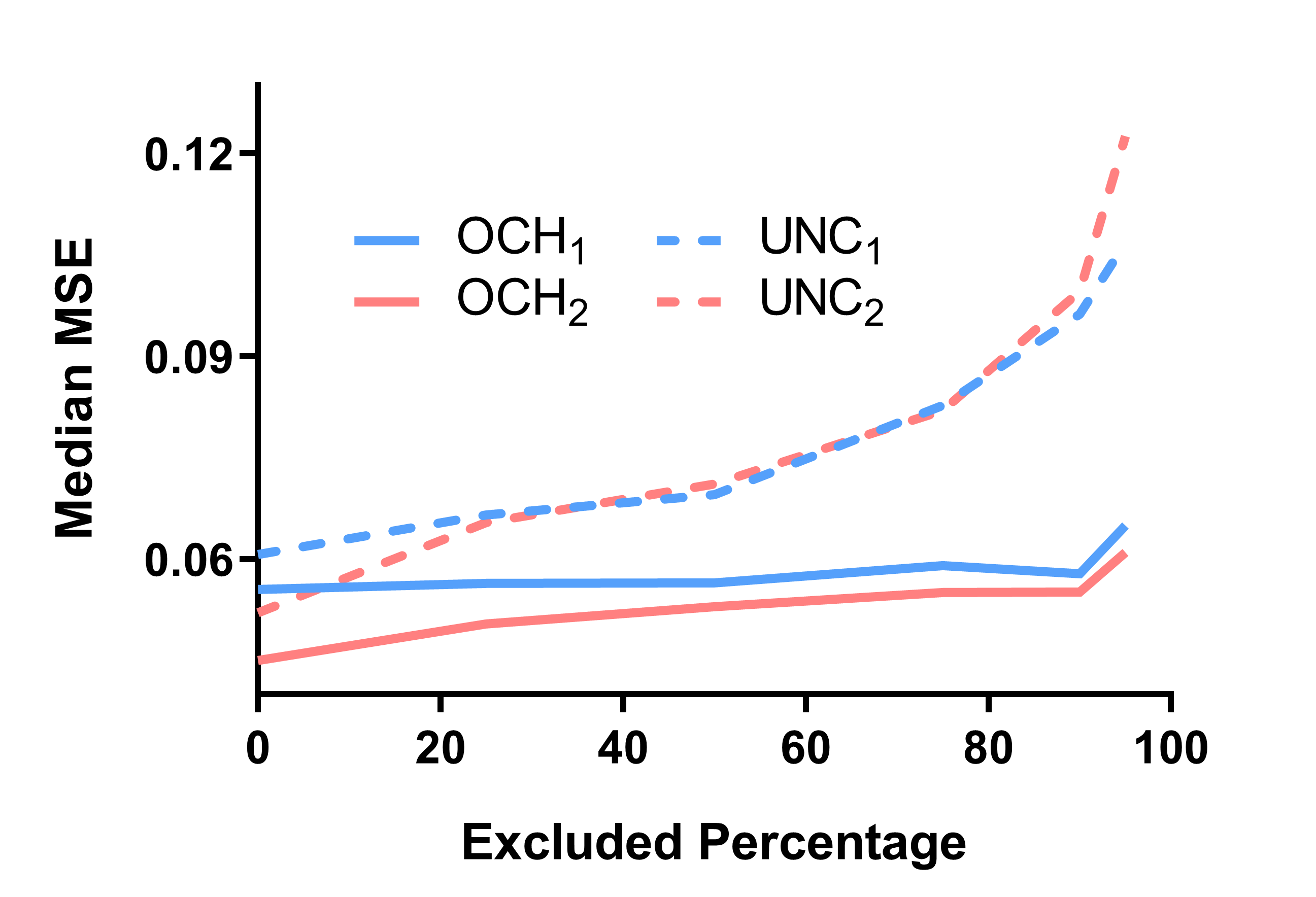}
    \caption{} \label{fig:median}
\end{subfigure}
\begin{subfigure}[]{0.5\textwidth}
\captionsetup{justification=centering,margin=2cm}
    \centering
    \includegraphics[scale=0.65]{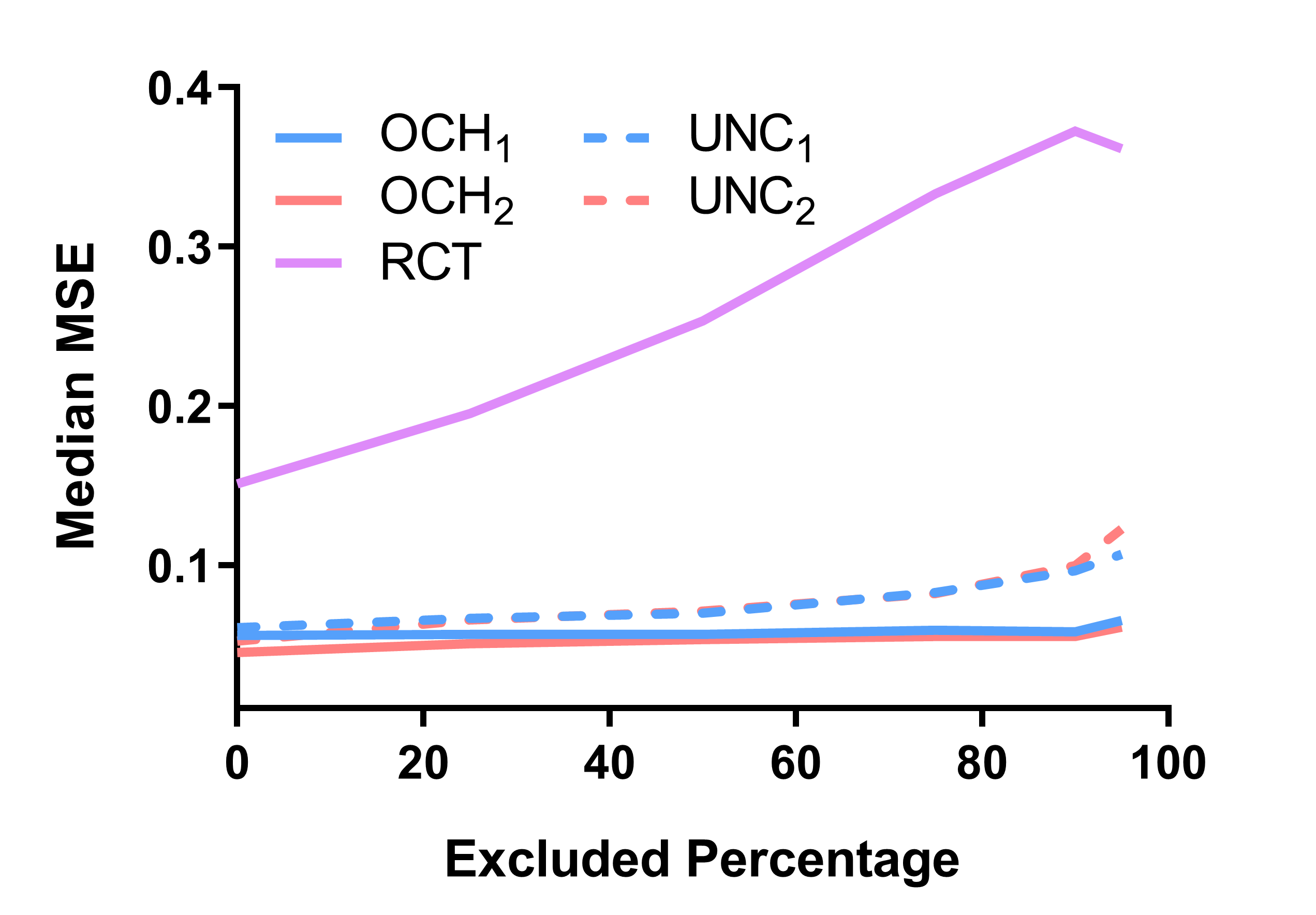}
    \caption{} \label{fig:medianRCT}
\end{subfigure}

\begin{subfigure}[]{0.5\textwidth}
\captionsetup{justification=centering,margin=2cm}
    \centering
    \includegraphics[scale=0.65]{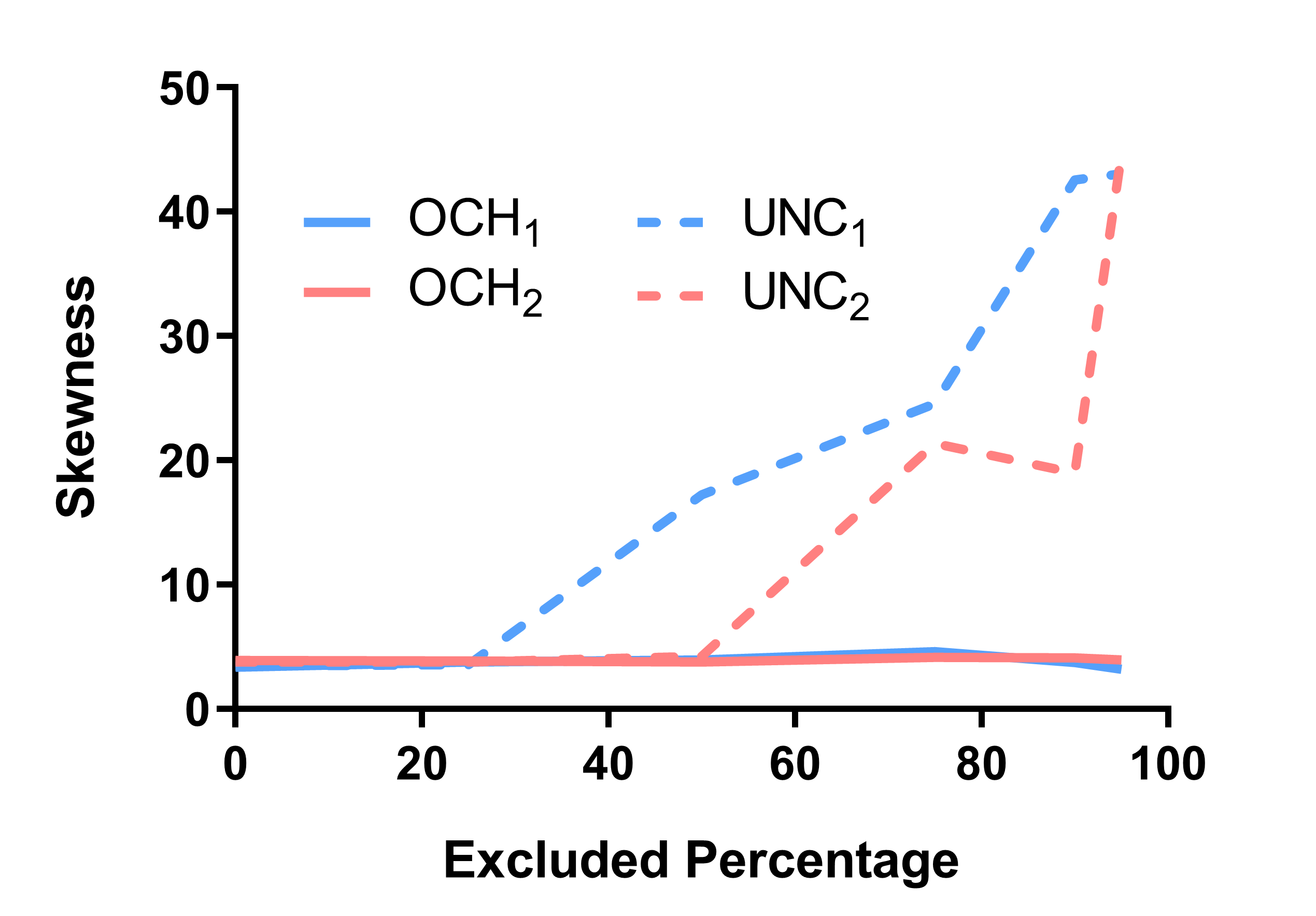}
    \caption{} \label{fig:skewness}
\end{subfigure}
\begin{subfigure}[]{0.5\textwidth}
\captionsetup{justification=centering,margin=2cm}
    \centering
    \includegraphics[scale=0.65]{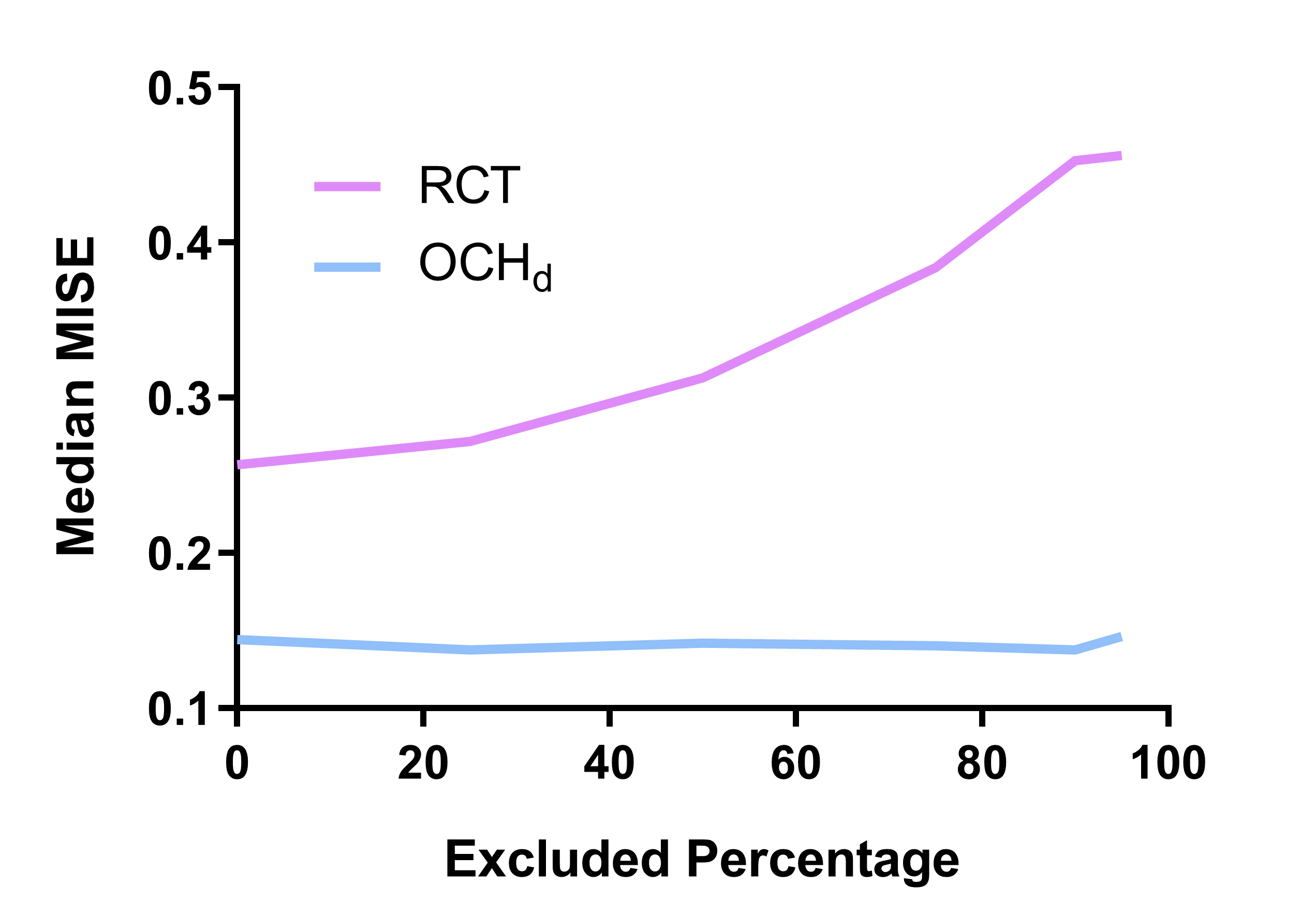}
    \caption{} \label{fig:OCHd}
\end{subfigure}
\caption{Stability results when Assumptions \ref{assump:main_e} and \ref{assump:main_p} hold. Lower and flatter is better. (a) The constraints in OCH$_2$ and OCH$_1$ prevent the MSE from growing with a higher percentage of excluded patients. (b) RCT offers terrible stability. (c) The constraints in OCH$_2$ and OCH$_1$ in particular prevent catastrophic failures by constraining skewness to the right (very high MSE values) in very exclusive trials. (d) OCH$_d$ also controls the MISE in the density setting, whereas RCT does not.} \label{fig:OCH}
\end{figure}

\noindent\textit{Stability.} Consistently good performance, i.e. stability across datasets, is important for high stakes areas like medicine. When Assumption \ref{assump:main_e} holds, OCH prevents the median MSE from growing even with the vast majority of patients excluded, while the UNC variants do not (Figure \ref{fig:median}). The deterioration in performance of RCT only is in fact much worse than even UNC$_2$ and UNC$_1$; the median MSE quickly increases with more stringent exclusion criteria (Figure \ref{fig:medianRCT}).

Closer inspection of the histograms show that the CATE OCH variants avoid catastrophic failures (very high MSE values) with higher percentages of excluded patients because the histogram of MSE values has minimal skewness to the right (to very large MSE values) (Figure \ref{fig:skewness}). In contrast, skewness quickly increases for the UNC variants when the majority of patients cannot enter the RCT. We conclude that the constraints in Expression \eqref{eq:optCATE_emp} are important for stability, particularly for very exclusive RCTs. Results when Assumption \ref{assump:main_e} is violated are qualitatively similar and presented in Appendix \ref{app:sim:viol}.

OCH$_d$ also prevents the MISE from growing as a larger proportion of patients are excluded from the RCT when Assumption \ref{assump:main_p} holds, unlike the RCT only algorithm; these results replicate those seen with the CATE (Figure \ref{fig:OCHd}). However, UNC$_d$ does not significantly increase skewness, again highlighting the minimal benefit of the constraints in the conditional density setting. Similar results apply when Assumption \ref{assump:main_p} is violated (Appendix \ref{app:sim:viol}).

\subsection{Real Data}

Evaluating the algorithms on real data is difficult because we rarely have access to the ground truth CATE or CDTE across the entire clinical population. Fortunately, investigators have conducted a handful of large, trans-institutional, multi-million dollar RCTs imposing few exclusion criteria. We use these RCTs to estimate the true CATE and CDTE. We then mimic more common exclusionary RCTs by imposing additional exclusion criteria. We finally generate observational data by asking a physician to remove patients who fail to match common prescribing patterns from the original RCTs. We present results for one real dataset here and refer the reader to Appendix \ref{app:real:add} for a second real dataset.

\begin{figure*}[t]
\begin{subfigure}[]{0.5\textwidth}
    \centering
    \captionsetup{justification=centering,margin=2cm}
    \includegraphics[scale=0.65]{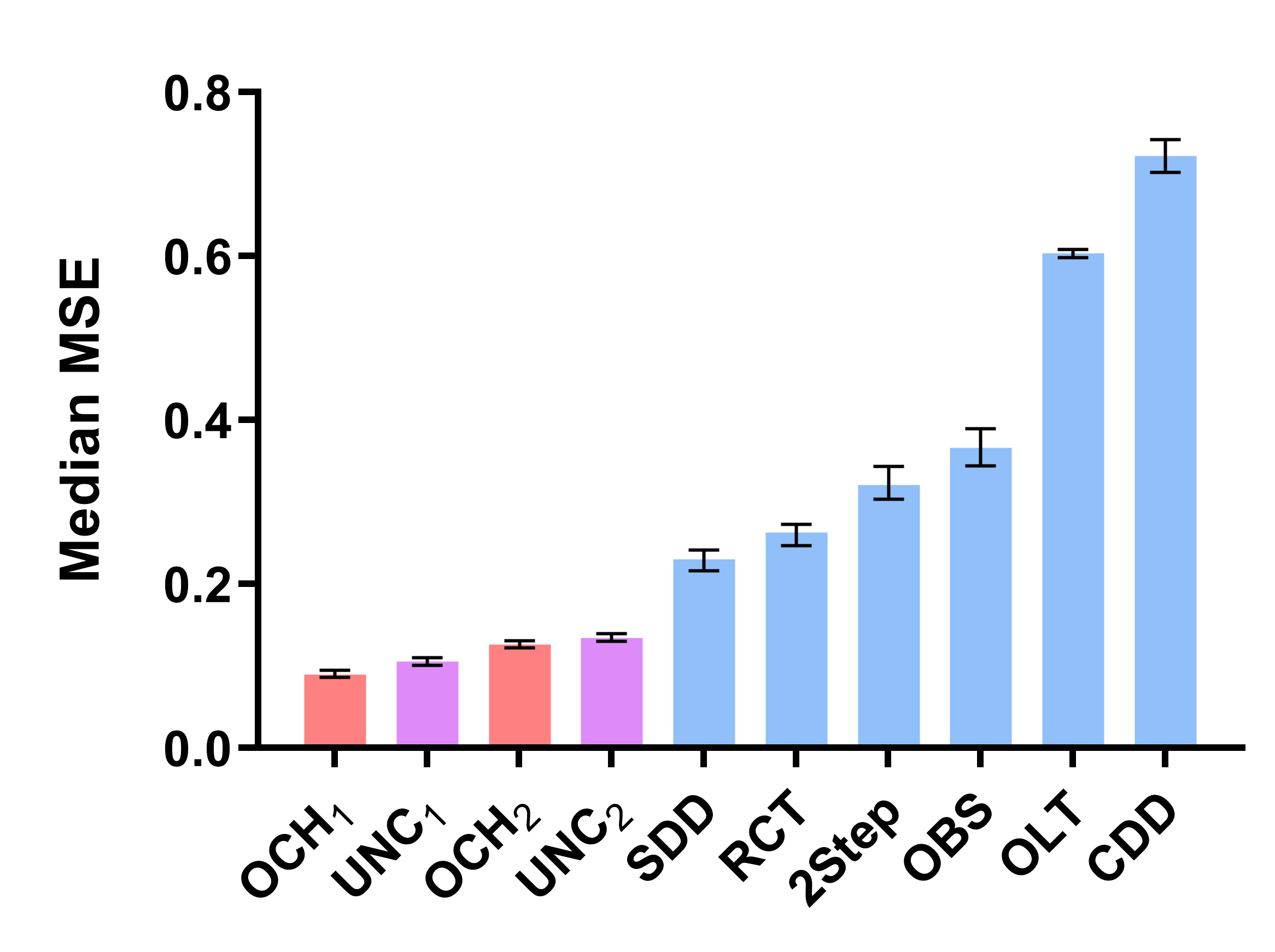}
    \caption{} \label{fig:STARD}
\end{subfigure}
\begin{subfigure}[]{0.5\textwidth}
    \centering
    \captionsetup{justification=centering,margin=2cm}
    \includegraphics[scale=0.65]{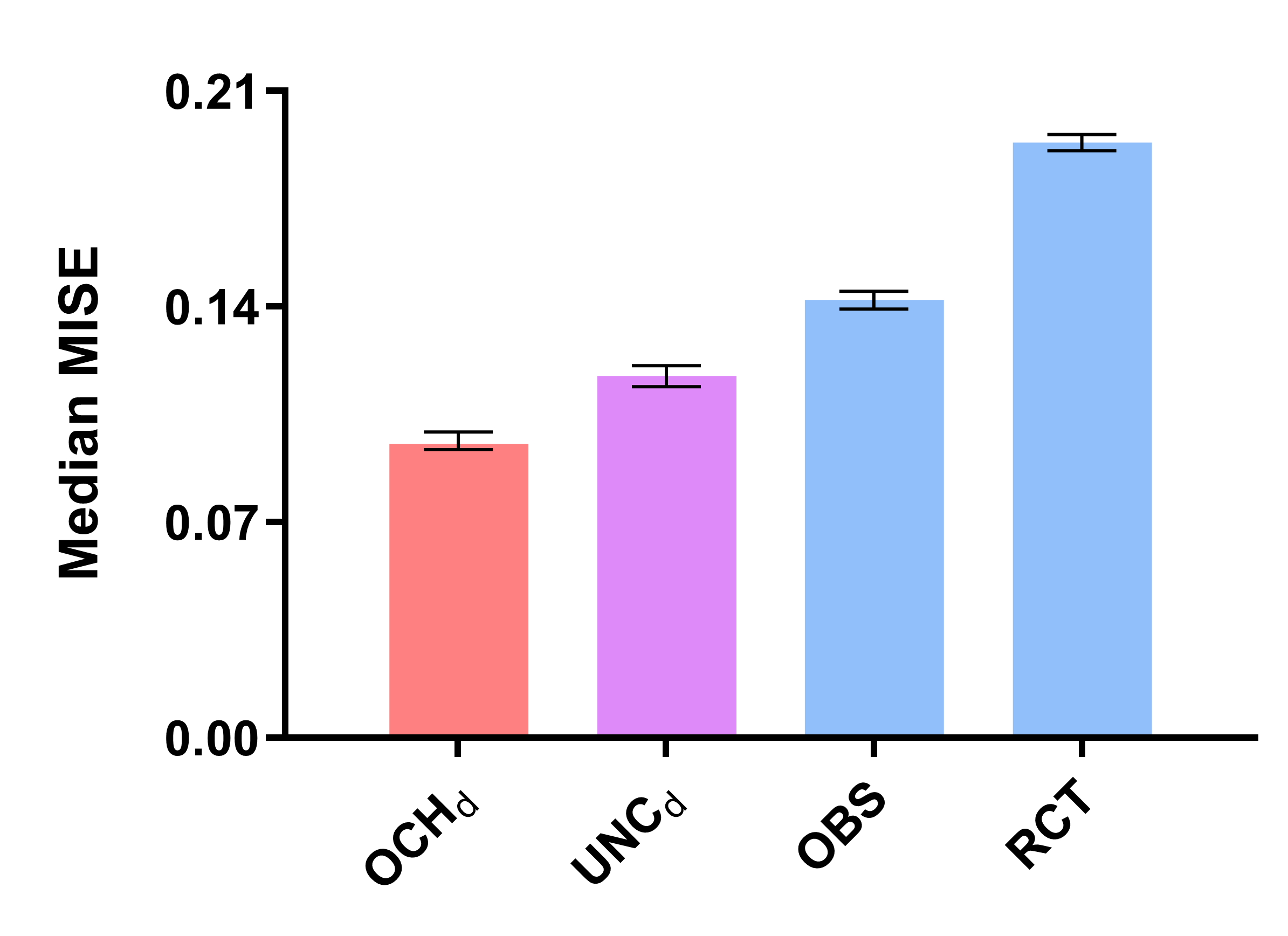}
    \caption{} \label{fig:STARD:dens}
\end{subfigure}
\caption{Real data results. We plot (a) median MSE values and (b) median MISE values for STAR*D. Error bars denote 95\% confidence intervals. The OCH algorithms in red usually achieve the best performance.} \label{fig:real}
\end{figure*}

Many randomized trials exclude patients who use other medications or illegal substances. Nevertheless, these patients are often the sickest.  We therefore evaluated the algorithms on how well they generalize to the broader population when trained on an RCT excluding this sub-group and a confounded observational dataset.

We obtained data from the Sequenced Treatment Alternatives to Relieve Depression trial (STAR*D), a large inclusive RCT designed to assess the sequential effects of anti-depressants and cognitive therapy on patients with major depressive disorder \citep{Rush06, Warden07,Trivedi06}. The investigators assessed treatment response using QIDS-SR, a self-reported measure of depressive symptoms. STAR*D ultimately included four levels of sequential treatment assignment.

We analyzed data from the second level because it had a large sample size and tested the effects of buproprion ($T=1$) versus venlafaxine and sertraline ($T=0$). Buproprion is known to have a unique effect on a symptom of depression called hypersomnia, or excessive sleepiness \citep{Papakostas06}. We examined the hypersomnia sub-score of QIDS-SR at week 6 in order to give sufficient time for the treatments to elicit differential effect. We used the other sub-scores in QIDS-SR related to sleep as predictors in $\bm{X}$, including
sleep onset insomnia, mid-nocturnal insomnia and early morning insomnia. We treated this dataset containing 388 samples as the comprehensive RCT.

We generated \textcolor{blue}{observational data} by imposing confounding on the comprehensive RCT. Physicians often prefer buproprion for patients with major depression who experience hypersomnia. We therefore removed patients receiving $T=1$ with no hypersomnia, or a hypersomnia QIDS-SR score of zero, but kept all patients receiving $T=0$. This process ultimately excluded 19.3\% of the original 388 patients. 

We next generated \textcolor{blue}{exclusive RCT data} by imposing additional exclusion criteria. We in particular performed a literature search and identified (1) current psychotropic use and (2) substance use as the two most common exclusion criteria in clinical trials of major depression not already implemented in STAR*D \citep{Blanco17}. We therefore excluded patients meeting at least one of those two criteria. This process eliminated 39.8\% of the original 388 patients.

We finally ran all of the algorithms on 2000 bootstrapped draws of the derived observational and exclusive RCT datasets. We quantify accuracy using either the MSE to the ground truth CATE, or the MISE to the ground truth CDTE estimated using all of the original 388 patients. For the CATE, both OCH$_2$ and OCH$_1$ outperform their predecessors (Figure \ref{fig:STARD}). The ablated variants UNC$_2$ and UNC$_1$ also perform well, but not as well as their constrained counterparts. The results for the CDTE are similar; OCH$_d$ performs the best, followed by UNC$_d$  (Figure \ref{fig:STARD:dens}). We conclude that all OCH algorithms perform well in estimating the CATE or CDTE even when including patients who use other psychotropics or substances (or both).

\section{Conclusion}

Physicians identify seemingly effective treatments by observing patient outcomes over time; they then readily administer those treatments to certain patients. We used this observation to propose a new approach to cross-design synthesis, where we bound the unconfounded treatment response between the confounded treatment response seen before and after treatment assignment.
This implies that the treatment effect must lie in the convex hull of two sets of conditional expectations or densities. We exploited the assumptions in three variants of the OCH algorithm which all analyze RCT and observational data simultaneously in order recover either the CATE or the CDTE over the entire population. Experimental results highlighted the superior performance of OCH compared to its predecessors. We conclude that OCH offers a promising new approach to generalizing randomized trials.

\bibliography{biblio}

\section{Appendix}
\subsection{Graphical Characterization} \label{app:graph}
We provide a graphical characterization of Assumptions \ref{assump:randomization}-\ref{assump:ignoreS} using Single World Intervention Graphs (SWIGs) \citep{Richardson13}. We prefer SWIGs over twin networks \citep{Balke94}, since SWIGs encode all conditional independence relations -- among the variables present in the SWIG -- that hold
for all distributions over counterfactuals. Twin networks lack this completeness property and were designed to model additional cross-world independencies, which we are not interested in in this paper. 

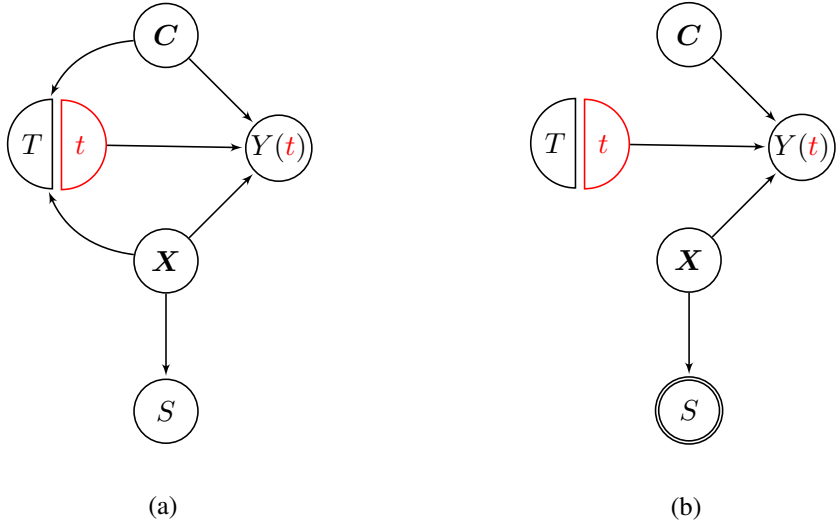
\begin{figure}
\begin{subfigure}{0.45\textwidth}
\begin{center}
\begin{tikzpicture}[scale=1.0, shorten >=1pt,auto,node distance=2.8cm, semithick,semic/.style args={#1,#2}{semicircle,minimum width=#1,draw,anchor=arc end,rotate=#2}]
                    
\tikzset{vertex/.style = {inner sep=0.4pt}}
\tikzset{edge/.style = {->,> = latex'}}
 
\node[semic={1cm,90}] (0) at  (0,-0.55) {\rotatebox{-90}{$T$}};
\node[semic={1.19cm,270},draw=red] (1) at  (0.1,0.65) {\rotatebox{90}{\textcolor{red}{$t$}}};
\node[vertex,draw=black, circle, minimum size=0.85cm] (2) at  (1.5,1.5) {$\bm{C}$};
\node[vertex,draw=black, circle] (3) at  (3,0) {$Y($\textcolor{red}{$t$}$)$};
\node[vertex,draw=black, circle, minimum size=0.85cm] (4) at  (1.5,-1.5) {$\bm{X}$};
\node[vertex,draw=black, circle, minimum size=0.85cm] (5) at  (1.5,-3.5) {$S$};

\draw[edge] (1) to (3);
\draw[edge,bend right=30] (2) to (0);
\draw[edge,bend left=30] (4) to (0);
\draw[edge] (4) to (5);
\draw[edge] (4) to (3);
\draw[edge] (2) to (3);
\end{tikzpicture}
\end{center}
\caption{} \label{fig_SWIG_OBS}
\end{subfigure}
\begin{subfigure}{0.45\textwidth}
\begin{center}
\begin{tikzpicture}[scale=1.0, shorten >=1pt,auto,node distance=2.8cm, semithick, semithick,semic/.style args={#1,#2}{semicircle,minimum width=#1,draw,anchor=arc end,rotate=#2}]
                    
\tikzset{vertex/.style = {inner sep=0.4pt}}
\tikzset{edge/.style = {->,> = latex'}}
 
\node[semic={1cm,90}] (1) at  (0,-0.55) {\rotatebox{-90}{$T$}};
\node[semic={1.19cm,270},draw=red] (1) at  (0.1,0.65) {\rotatebox{90}{\textcolor{red}{$t$}}};
\node[vertex,draw=black, circle, minimum size=0.85cm] (2) at  (1.5,1.5) {$\bm{C}$};
\node[vertex,draw=black, circle] (3) at  (3,0) {$Y($\textcolor{red}{$t$}$)$};
\node[vertex,draw=black, circle, minimum size=0.85cm] (4) at  (1.5,-1.5) {$\bm{X}$};
\node[vertex,draw=black, circle, minimum size=0.85cm,accepting] (5) at  (1.5,-3.5) {$S$};

\draw[edge] (1) to (3);
\draw[edge] (4) to (5);
\draw[edge] (4) to (3);
\draw[edge] (2) to (3);
\end{tikzpicture}
\end{center}
\caption{} \label{fig_SWIG_RCT}
\end{subfigure}
\caption{SWIGs for the observational distribution in (a) and the RCT distribution in (b).} \label{fig_SWIG}
\end{figure}

We first require some definitions. A \textit{directed} \textit{graph} $\mathbb{G}$ is a graph over a set of vertices $\bm{Z}$ with at most one directed edge between any two vertices. A \textit{directed} \textit{path} from $A$ to $B$ is a sequence of directed edges from $A$ to $B$. A \textit{cycle} occurs when there exists a directed path from $A$ to $B$ and $B \rightarrow A$. A \textit{directed} \textit{acyclic} \textit{graph} (DAG) is a directed graph without cycles. $A$ is an \textit{ancestor} of $B$ if there exists a directed path from $A$ to $B$. A \textit{collider} refers to the triple $A \rightarrow B \leftarrow C$. Two vertices $A$ and $B$ are \textit{d-connected} given $\bm{W} \subseteq (\bm{Z} \setminus \{A,B\})$ in a DAG, if there exists a path between $A$ and $B$ such that any collider on the path is an ancestor of $\bm{W}$ and no non-collider on the path is in $\bm{W}$. Otherwise, $A$ and $B$ are \textit{d-separated} given $\bm{W}$.

We summarize the causal relations using the SWIG of the observational distribution in Figure \ref{fig_SWIG_OBS}; a SWIG is a DAG  where we split the treatment vertex $T$ into $T$ and $t$ and replace the outcome $Y$ with the counterfactual variable $Y(t)$. The vertex $\bm{C}$ denotes a set of potentially unobserved confounders. We obtain the SWIG of the RCT distribution in Figure \ref{fig_SWIG_RCT} by removing the directed edges into $T$. The d-separation relation between $Y(t)$ and $T$ given $(\bm{X},S)$ in Figure \ref{fig_SWIG_RCT}, and likewise the d-separation relation between $\bm{X}$ and $T$ given $S$ imply Assumption \ref{assump:randomization}. Assumption \ref{assump:ignoreS} corresponds to the d-separation relation between $Y(t)$ and $S$ given $\bm{X}$. The second assumption holds from the fact that $S=1$ in the RCT distribution but can take on any value in the observational distribution.

\newpage
\subsection{CATE versus CDTE} \label{app:CATEvCDTE}

\subsection{Counter-Example} \label{app:CE}

 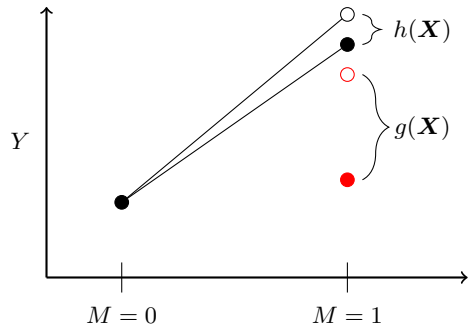
\begin{figure}[h]
\begin{center}
\begin{tikzpicture}[mydrawstyle/.style={draw=black, very thick}, x=1mm, y=1mm, z=1mm]

  \draw[thick, ->](0,30)--(0,66);
  \draw[thick, ->](0,30)--(56,30);

  \draw (10,28)--(10,32);
  \draw (40,28)--(40,32);

  \node[] at (10,25) {\smaller $M=0$};
  \node[] at (40,25) {\smaller $M=1$};

   \node[rotate=0] at (-3.5,48){\begin{tabular}{c} \smaller{$Y$} \end{tabular}};

   \node[circle,draw,fill=black,scale=0.5] (a) at (10,40){}; 
   \node[circle,draw,fill=black,scale=0.5] (b) at (40,61){};
   \draw (a) -- (b);
 
   \node[circle,draw,scale=0.5] (c) at (40,65){};
   \draw (a) -- (c);

      \node[circle,draw=red,fill=red,scale=0.5] (c) at (40,43){};
         \node[circle,draw=red,scale=0.5] (c) at (40,57){};

     \draw [decorate,
    decoration = {brace,amplitude=10pt}] (42,57)--(42,43);
    \node[rotate=0] at (50,50){\begin{tabular}{c} \smaller{$g(\bm{X})$}\end{tabular}};

         \draw [decorate,
    decoration = {brace,amplitude=7pt}] (42,65)--(42,61);
    \node[rotate=0] at (50,63){\begin{tabular}{c} \smaller{$h(\bm{X})$}\end{tabular}};

\end{tikzpicture}
\end{center}
\caption{Example of a situation where $g(\bm{X})$ is larger than $h(\bm{X})=\mathbb{E}^o(Y_1|\bm{X},T=1)-\mathbb{E}^o(Y_1|\bm{X},T=0)$ but Assumption \ref{assump:main_e} holds. The two black lines correspond to the change in expected outcomes of the two treatments in the observational distribution, while the two red dots correspond to the expected outcomes of the two treatments in the RCT distribution at $M=1$.} \label{fig:ex:bigger}

\end{figure}

\subsection{Conditional Densities of Treatment Effect} \label{app:CDTE}

\subsubsection{CDTE with Two Time Steps}

The CATE only provides a point estimate of the differential effect of treatment. As mentioned previously, we would like to visualize the probabilities associated with all possible outcome values using the CDTE. Assumption \ref{assump:main_p} helps us recover the CDTE using two time steps. The assumption requires $p^o(Y_1|\bm{X},T)$ and $p^o(Y_0|\bm{X})$, which we can obtain from the observational distribution. The mixing proportion $\mu_T \in [0,1]$ remains unspecified, but we have $p^r(Y_1(T)|\bm{X},T,S=1) =  p^r(Y_1(T)|\bm{X},S=1) =  p^r(Y_1(T)|\bm{X})$ in the RCT distribution by Assumptions \ref{assump:randomization} and \ref{assump:ignoreS}, respectively. We can therefore solve for the optimal $\mu_T^*$ by minimizing the following mean integrated squared error (MISE) quantifying the distance between the mixture density $p^o(Y_1|\bm{X},T)\mu_T + p^o(Y_0|\bm{X})(1-\mu_T)$ and the desired density $p^r(Y_1(T)|\bm{X})$:
\begin{equation} \label{eq:optCDTE}
\begin{aligned}
    \mu_T^* = \argmin_{\mu_T} \mathbb{E}^r_{\bm{X}|S=1}\Big[\int \big( [p^o(Y_1=y|\bm{X},T)\mu_T + &p^o(Y_0=y|\bm{X})(1-\mu_T)]\\ &- p^r(Y_1(T)=y|\bm{X}) \big)^2 ~ dy \Big]\\
     &\hspace{-35mm} s.t. \hspace{2mm} 0 \leq \mu_T \leq 1,
\end{aligned}
\end{equation}
\noindent where the expectation is taken with respect to the RCT distribution, or on $\mathcal{S}_R$. We can solve Expression \eqref{eq:optCDTE} without access to \textcolor{blue}{$p^r(Y_1(T)|\bm{X})$} because the objective function is proportional to:
\begin{equation}  \label{eq:optCDTE_expand}
\begin{aligned}
    &\hspace{10mm}\frac{1}{2} \mathbb{E}^r_{\bm{X}|S=1} \Big[\int p^o(Y_1|\bm{X},T)^2 + p^o(Y_0|\bm{X})^2 - 2p^o(Y_1|\bm{X},T)p^o(Y_0|\bm{X}) ~dy\Big]\mu_T^2 - \\ \Big[&\underbrace{\mathbb{E}^r_{Y_1(T),\bm{X}|S=1} \Big( p^o(Y_1|\bm{X},T) - p^o(Y_0|\bm{X}) \Big)}_{\textnormal{Expectation w.r.t. $\textcolor{blue}{\mathbb{P}^r(Y_1(T)|\bm{X})}\mathbb{P}^r(\bm{X}|S=1)$}} - \mathbb{E}^r_{\bm{X}|S=1} \Big(\int p^o(Y_1|\bm{X},T)p^o(Y_0|\bm{X}) - p^o(Y_0|\bm{X})^2\Big) \Big]\mu_T,
\end{aligned}
\end{equation}
obtained by expanding the square and dropping constants that do not depend on $\mu_T$. This new form does not require access to $p^r(Y_1(T)|\bm{X})$, but only to the expectation with respect to $\mathbb{P}^r(Y_1(T)|\bm{X},S=1)\mathbb{P}^r(\bm{X}|S=1) = \mathbb{P}^r(Y_1(T)|\bm{X})\mathbb{P}^r(\bm{X}|S=1)$ as indicated by the underbrace per Assumption \ref{assump:ignoreS}. We therefore obtain the same $\mu_T^*$ by replacing the objective function in Expression \eqref{eq:optCDTE} with the one above.

We have the result below to seal the strategy by following a similar argument as Theorem \ref{thm:CATE} but extended to conditional densities:
\begin{theorem1}
The quantity $p^o(Y_1|\bm{X},T)\mu_T^* + p^o(Y_0|\bm{X})(1-\mu_T^*)$ is equivalent to $p^r(Y_1(T)|\bm{X})$ on $\mathcal{S}_O$ under Assumptions \ref{assump:randomization}-\ref{assump:ignoreS} and \ref{assump:main_p}.
\end{theorem1}
\begin{proof}
$p^r(Y_1(T)|\bm{X})$ is equivalent to $p^o(Y_1|\bm{X},T)\mu_T + p^o(Y_0|\bm{X})(1-\mu_T)$ for some $\mu_T \in [0,1]$ by Assumption \ref{assump:main_p}. We also know that $p^r(Y_1(T)|\bm{X},T,S=1)$ is equivalent to $p^r(Y_1(T)|\bm{X})$ in the RCT distribution under Assumptions \ref{assump:randomization} and \ref{assump:ignoreS}: $p^r(Y_1(T)|\bm{X},T,S=1)\stackrel{1}{=}  p^r(Y_1(T)|\bm{X},S=1) \stackrel{3}{=}  p^r(Y_1(T)|\bm{X})$. The quantity $p^o(Y_1|\bm{X},T)\mu_T + p^o(Y_0|\bm{X})(1-\mu_T)$ is unique on $\mathcal{S}_O$ for any $\mu_T \in \mathbb{R}$. The solution $\mu_T^*$ to Expression \eqref{eq:optCDTE} is unique on $\mathcal{S}_R$ because $\mathcal{S}_O \cap \mathcal{S}_R = \mathcal{S}_R$ by Assumption \ref{assump:support}. The quantity $p^o(Y_1|\bm{X},T)\mu_T^* + p^o(Y_0|\bm{X})(1-\mu_T^*)$ is therefore equivalent to $p^r(Y_1(T)|\bm{X})$ on $\mathcal{S}_O$.
\end{proof}

We present the corresponding algorithm called Optimum in Convex Hulls for Densities (OCH$_d$) for the finite sample setting in Algorithm \ref{alg:OCH$_d$}. The method is similar to Algorithm \ref{alg:OCH2} with some important differences. First, OCH$_d$ does not directly approximate the CDTE using the RCT. The algorithm instead approximates $p^o(Y_0|\bm{X})$ in Step \ref{alg:OCH$_d$:Y_0} and $p^o(Y_1|\bm{X},T)$ in Step \ref{alg:OCH$_d$:Y_1(1)} using the observational data. OCH$_d$ then obtains the empirical estimate of $\mu_T^*$ in Step \ref{alg:OCH$_d$:mu} by solving the following quadratic program -- equivalent to the empirical version of Expression \eqref{eq:optCDTE_expand}, obtained by replacing expectations with means and densities with their estimates:
\begin{equation} \label{eq:optCDTE_emp}
\begin{aligned}
\widehat{\mu}_T &= \argmin_{\mu_T} \hspace{2mm} \frac{1}{2}H \mu_T^2 - d\mu_T\\
    &\hspace{10mm} s.t. \hspace{2mm} 0 \leq \mu_T \leq 1,
\end{aligned}
\end{equation}
where:
\begin{equation*}
\begin{aligned}
    &H = \frac{1}{2n} \sum_{i=1}^{2n} \Big[\int \widehat{p}^o(Y_1=y|\bm{x}_i,T)^2 + \widehat{p}^o(Y_0=y|\bm{x}_i)^2 - 2\widehat{p}^o(Y_1=y|\bm{x}_i,T)\widehat{p}^o(Y_0=y|\bm{x}_i) ~dy\Big]\\
&d = \frac{1}{n}\sum_{i=1}^{n} \Big(\widehat{p}^o(Y_1=y_{1i}(T)|\bm{x}_i)-\widehat{p}^o(Y_0=y_{1i}(T)|\bm{x}_i)\Big)\\ &\hspace{40mm}-\frac{1}{2n}\sum_{i=1}^{2n} \Big(\int \widehat{p}^o(Y_1=y|\bm{x}_i,T)\widehat{p}^o(Y_0=y|\bm{x}_i) - \widehat{p}^o(Y_0=y|\bm{x}_i)^2 ~dy\Big).
\end{aligned}
\end{equation*}
Finally, the algorithm predicts $\widehat{p}^o(Y_1|\bm{x},T)\widehat{\mu}_T + \widehat{p}^o(Y_0|\bm{x})(1-\widehat{\mu}_T)$ for all $\bm{x}$ in the test set $\mathcal{T}$ each lying anywhere in $\mathcal{S}_O$.

\begin{algorithm}[]
 \nonl \textbf{Input:} trial data, observational data, test points $\mathcal{T}$\\
 \nonl \textbf{Output:} $\widehat{p}^o(Y_1|\bm{X},t)\widehat{\mu}_t + \widehat{p}^o(Y_0|\bm{X})(1-\widehat{\mu}_t)$ on $\mathcal{T}$ for $t \in \{0,1\}$\\
 \BlankLine

Estimate $p^o(Y_0|\bm{X})$ \label{alg:OCH$_d$:Y_0}\\
\For{$t \in \{0,1\}$}{
Estimate $p^o(Y_1|\bm{X},t)$ on $\mathcal{S}_O$ using the observational data \label{alg:OCH$_d$:Y_1(1)}\\
Solve Expression \eqref{eq:optCDTE_emp} using $\widehat{p}^o(Y_1|\bm{X},t)$ and $\widehat{p}^o(Y_0|\bm{X})$ on the trial data \label{alg:OCH$_d$:mu}\\
Predict $\widehat{p}^o(Y_1|\bm{X},t)\widehat{\mu}_t + \widehat{p}^o(Y_0|\bm{X})(1-\widehat{\mu}_t)$ on $\mathcal{T}$ \label{alg:OCH$_d$:test}
}

 \caption{Optimum in Convex Hulls for Densities (OCH$_d$)} \label{alg:OCH$_d$}
\end{algorithm}

\subsubsection{CDTE with One Time Step}

Estimating the CDTE with one time step is unfortunately not a straightforward modification of the CDTE with two time steps as with the CATE. The problem stems from the indeterminancy of a ``worst case scenario'' for $p^o(Y_0|\bm{X})$. Simply choosing the Dirac delta function at zero does not work because, if $p^r(Y_1(T)|\bm{X})$ is contained in the convex hull of $p^o(Y_1|\bm{X},T)$ and $p^o(Y_0|\bm{X})$, then $p^r(Y_1(T)|\bm{X})$ is \textit{not} necessarily contained in the convex hull of $p^o(Y_1|\bm{X},T)$ and the Dirac delta. In contrast, if $\mathbb{E}^r(Y_1(T)|\bm{X})$ is contained in the convex hull of $\mathbb{E}^o(Y_1|\bm{X},T)$ and $\mathbb{E}^o(Y_0|\bm{X})$, then $\mathbb{E}^r(Y_1(T)|\bm{X})$ is clearly contained in the convex hull of $E^o(Y_1|\bm{X},T)$ and 0 because $\mathbb{E}^o(Y_0|\bm{X}) \geq 0$ with suitable normalization. The logic with conditional expectations therefore does not carry over to conditional densities. Estimating the CDTE with one time step likely requires a non-linear generalization of the convex hull, so we leave it open to future work.

\subsection{Related Work} \label{app:RW}

The variants of OCH fall into a category of methods that accomplish \textit{cross-design synthesis}, \textit{transportability}, \textit{data fusion} or \textit{generalizability}; these terms refer to the act of combining trial and observational data (and potentially other dataset types) in order to the eliminate the weaknesses of each \citep{Droitcour93,Stuart15,Bareinboim16}. Note that a plethora of algorithms, such as inverse probability weighted estimators, exist when $\mathcal{S}_O \subseteq \mathcal{S}_R$, but relatively few methods investigate the more realistic situation when $\mathcal{S}_R \subset \mathcal{S}_O$ with \textit{strict} exclusion criteria in randomized trials \citep{Rosenbaum83,Lunceford04}. The earliest methods in the latter category assume unconfoundedness and simply estimate the CATE using the observational data \citep{Rosenbaum83,Pearl09}. The weakness of these methods of course lies in the unconfoundedness assumption, which we can neither guarantee nor verify in practice.

Later algorithms proposed to modify the observational estimate $f(\bm{X}) = \mathbb{E}^o(Y_1|\bm{X},T=1) - \mathbb{E}^o(Y_1|\bm{X},T=0)$ using a linear transformation. The earliest algorithm in this category, which we call Outer Linear Transform (OLT), proposed:
\begin{equation*}
    g(\bm{X}) = f(\bm{X})\alpha + \beta,
\end{equation*}
where $\alpha$ and $\beta$ are fit using linear regression with trial data \citep{Jackson17}. OLT has the desirable property of preserving the shape of $f(\bm{X})$, but it offers limited flexibility in adjusting $f(\bm{X})$. A subsequent method called 2Step proposed to modify $f(\bm{X})$ using a linear combination of the predictors \citep{Kallus18}:
\begin{equation*}
    g(\bm{X}) = f(\bm{X}) + \bm{X}\delta + \beta.
\end{equation*}
It remains unclear however \textit{why} $f(\bm{X})$ should be linearly related to $g(\bm{X})$ via $\bm{X}$. Subsequent authors therefore proposed a more principled approach to choosing the basis functions in an algorithm called Synthesized Difference in Differences (SDD) \citep{Strobl21}. SDD linearly combines four conditional expectations as follows:
\begin{equation*}
\begin{aligned}
    g(\bm{X}) = &[\mathbb{E}^o(Y_1|\bm{X},T=1) - \mathbb{E}^o(Y_0|\bm{X},T=1)\alpha_1]\\ &- [\mathbb{E}^o(Y_1|\bm{X},T=0)\alpha_2 - \mathbb{E}^o(Y_0|\bm{X},T=0)\alpha_3],
\end{aligned}
\end{equation*}
The authors showed that this transformation relaxes the parallel slopes assumption used in the conditional Difference in Differences algorithm \citep{Abadie05}. SDD therefore carries theoretical justification, but it performs unstably in practice; the algorithm requires a substantial amount of regularization in order to consistently estimate the CATE with a high degree of accuracy, calling into question whether the underlying assumption actually holds in practice.

The OCH variants synthesize RCT and observational data by exploiting fundamentally different assumptions than those adopted by prior methods -- Assumptions \ref{assump:main_e} or \ref{assump:main_p}. The OCH algorithms in particular assume that confounding may exacerbate the magnitude of the treatment response over time, but it preserves the direction; this weakens the unconfoundedness assumption which requires that both the magnitude and direction are preserved once we condition on $\bm{X}$. OCH thus provides a theoretical explanation as to \textit{why} the adopted basis functions should be linearly adjusted in order to recover the CATE. OCH also introduces regularization naturally into Expressions \eqref{eq:optCATE} and \eqref{eq:optCDTE} via the convex hull. Finally, the algorithms extend the CATE and the CDTE to $\mathcal{S}_O$, whereas prior methods only extend the CATE. OCH therefore offers a superior approach to cross-design synthesis compared to its predecessors.

More generally, OCH capitalizes on a mixture of distributions which has already been exploited in causal graph learning with large gains in performance relative to methods that assume a single distribution \citep{Strobl19,Strobl22}. OCH was originally inspired by this mixture framework, even though the variants tackle a different problem. If accounting for mixtures improves performance with causal graph learning, then it should also improve performance with treatment effect estimation.  

\newpage
\subsection{Additional Experimental Results when Assumptions Hold} \label{app:sim:add}

\begin{table}[h]
\centering
\begin{subtable}{0.98\textwidth}  
\centering
\begin{tabular}{ccccc}
\hhline{=====}
      & 1               & 3               & 6               & 10              \\  \hline
\rowcolor{Gray} OCH$_2$   & \textbf{0.0186} & \textbf{0.0353} & \textbf{0.0703} & \textbf{0.1170} \\
\rowcolor{Gray} OCH$_1$   & \textbf{0.0214}         & \textbf{0.0439}          & \textbf{0.0824}          & \textbf{0.1302}          \\
\rowcolor{Gray} UNC$_2$   & 0.0637         & 0.0465         & 0.0821 & 0.1318\\
\rowcolor{Gray} UNC$_1$   & 0.0517          & 0.0485          & 0.0914 & 0.1458\\
SDD   & 0.0627          & 0.0962          & 0.1666          & 0.2336          \\
2Step & 0.0742          & 0.1767          & 0.3320          & 0.4987          \\
RCT   & 0.1525          & 0.1735          & 0.2500          & 0.4009          \\
OBS   & 0.0628          & 0.1545          & 0.3155          & 0.5248         \\
OLT   & 0.3314          & 0.3922          & 0.4908          & 0.6732          \\
CDD   & 0.2551          & 0.4681          & 0.8057          & 1.2309          \\  
\hhline{=====}
\end{tabular}
\caption{} \label{exp:CATE:vars}
\end{subtable}

\vspace{3mm}\begin{subtable}{0.98\textwidth}  
\centering
\captionsetup{justification=centering,margin=2cm}
\begin{tabular}{ccccc}
\hhline{=====}
      & 1               & 3               & 6               & 10              \\  \hline
\rowcolor{Gray} OCH$_d$   & \textbf{0.0272} & \textbf{0.1157} & \textbf{0.1906} & \textbf{0.2425} \\
\rowcolor{Gray} UNC$_d$   & 0.0319          & \textbf{0.1166}          & \textbf{0.1914} & \textbf{0.2428}\\
OBS   & 0.1042          & 0.2024          & 0.2693          & 0.3072          \\
RCT   & 0.2895          & 0.3191          & 0.3719          & 0.3703         \\
\hhline{=====}
\end{tabular}
\caption{} \label{table:CDTE:vars}
\end{subtable}
\caption{Accuracy results for (a) the CATE and (b) the CDTE across all numbers of variables in $\bm{X}$ when Assumptions \ref{assump:main_e} and \ref{assump:main_p} hold.} \label{exp:add:vars}
\end{table}

\newpage
\subsection{Experimental Results when Assumptions are Violated} \label{app:sim:viol}

\begin{figure}[h] \label{fig:OCHv}
\begin{subfigure}[]{0.5\textwidth}
\captionsetup{justification=centering,margin=2cm}
    \centering
    \includegraphics[scale=0.65]{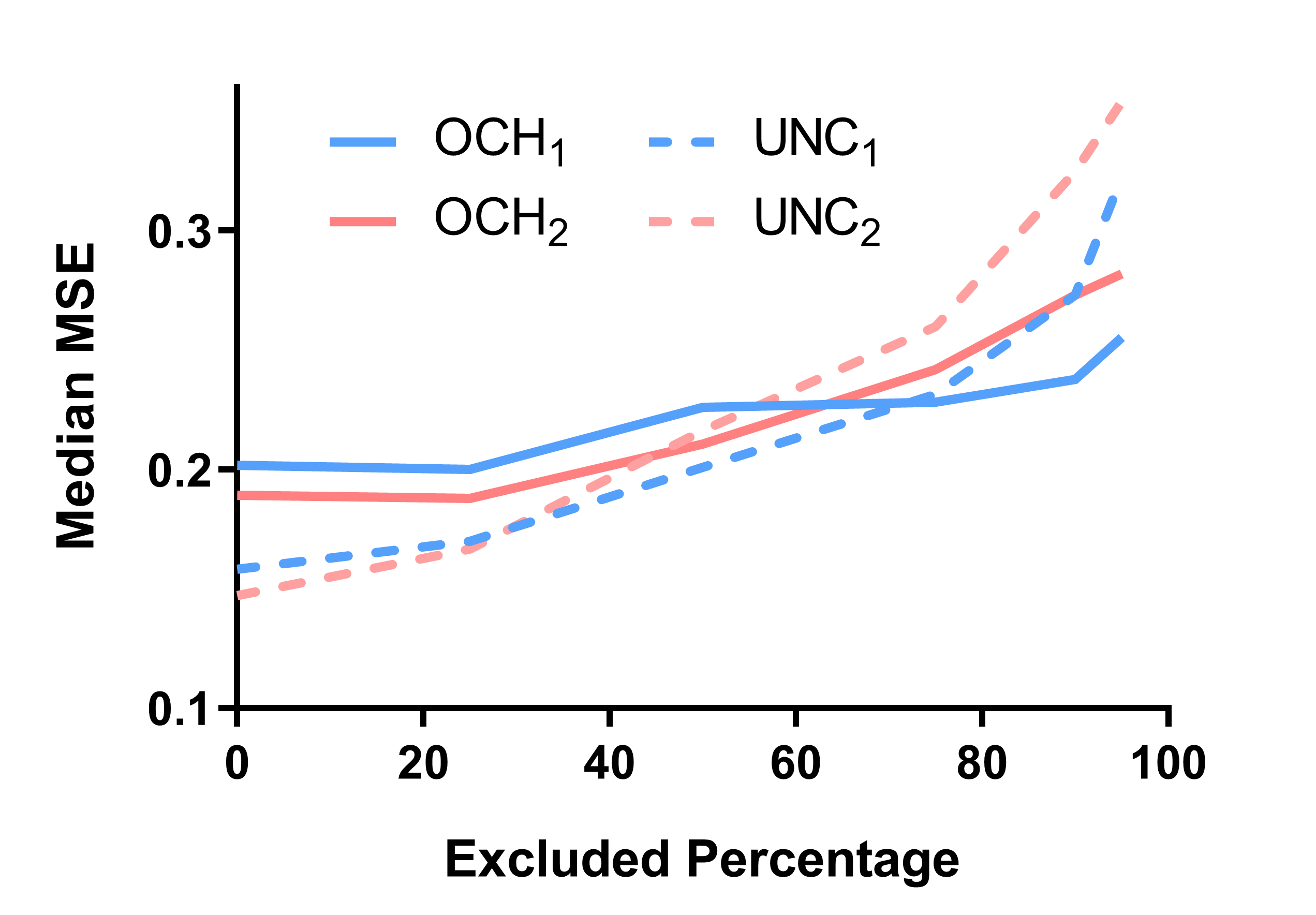}
    \caption{} \label{fig:medianv}
\end{subfigure}
\begin{subfigure}[]{0.5\textwidth}
\captionsetup{justification=centering,margin=2cm}
    \centering
    \includegraphics[scale=0.65]{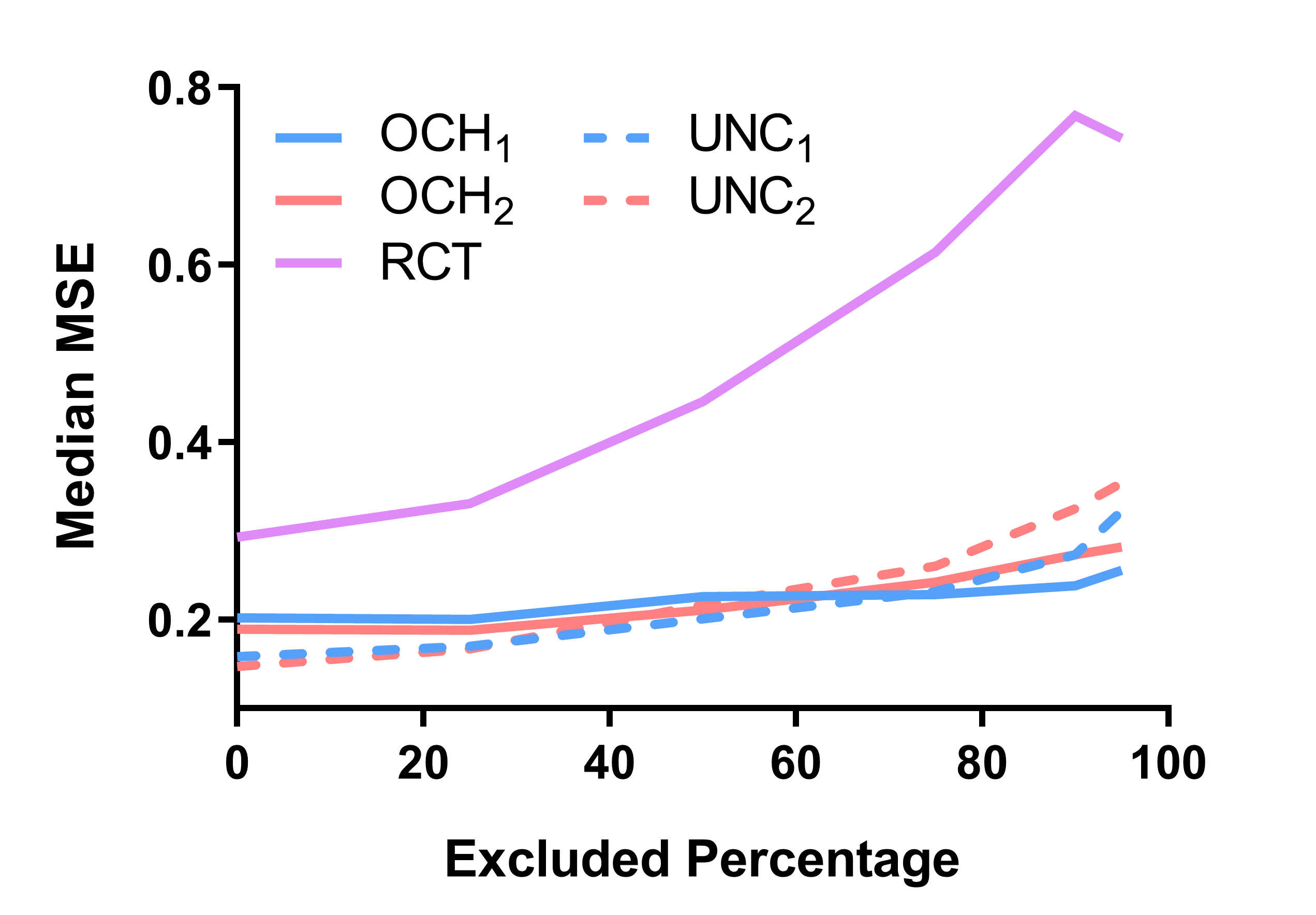}
    \caption{} \label{fig:medianRCTv}
\end{subfigure}

\begin{subfigure}[]{0.5\textwidth}
\captionsetup{justification=centering,margin=2cm}
    \centering
    \includegraphics[scale=0.65]{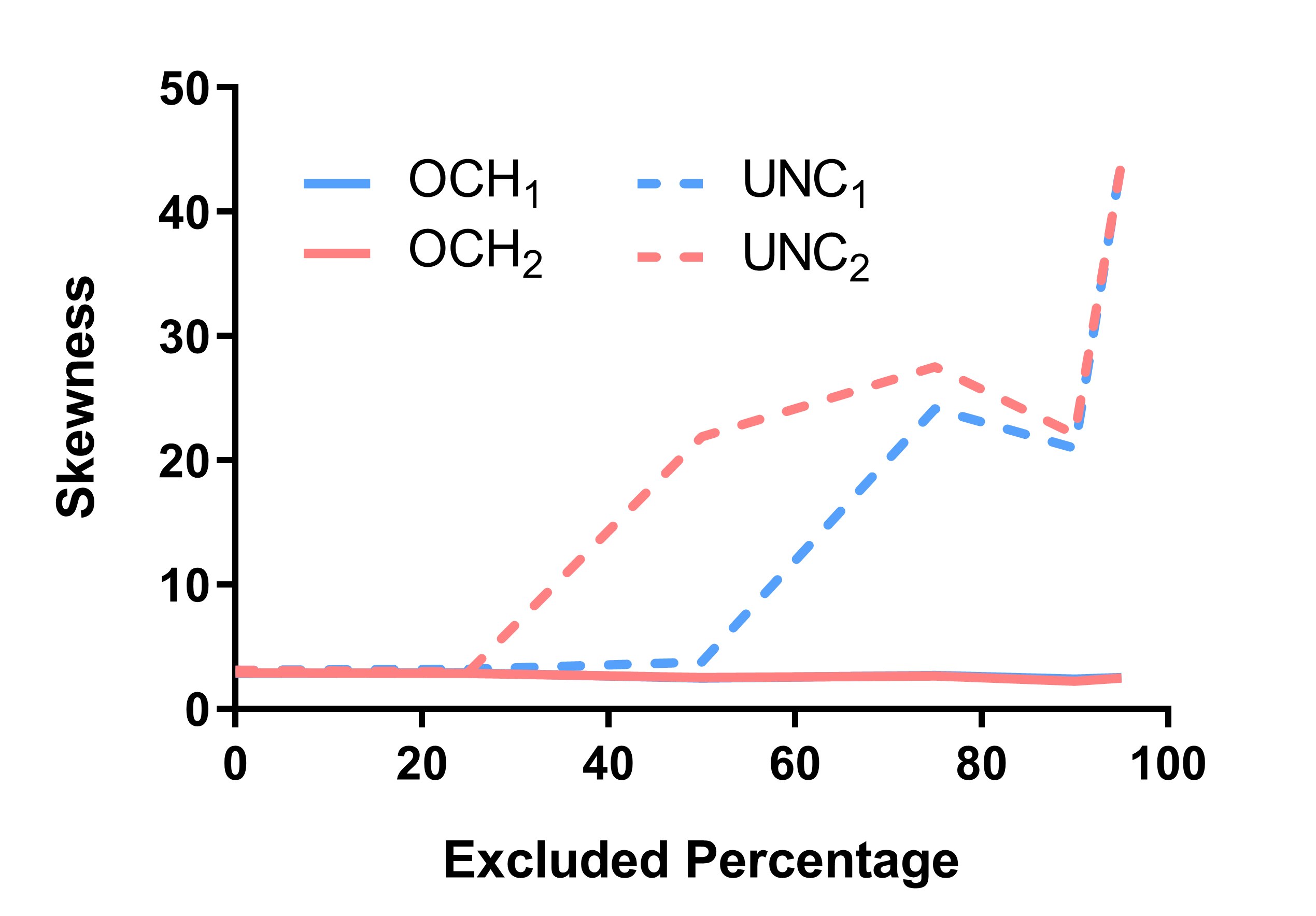}
    \caption{} \label{fig:skewnessv}
\end{subfigure}
\begin{subfigure}[]{0.5\textwidth}
\captionsetup{justification=centering,margin=2cm}
    \centering
    \includegraphics[scale=0.65]{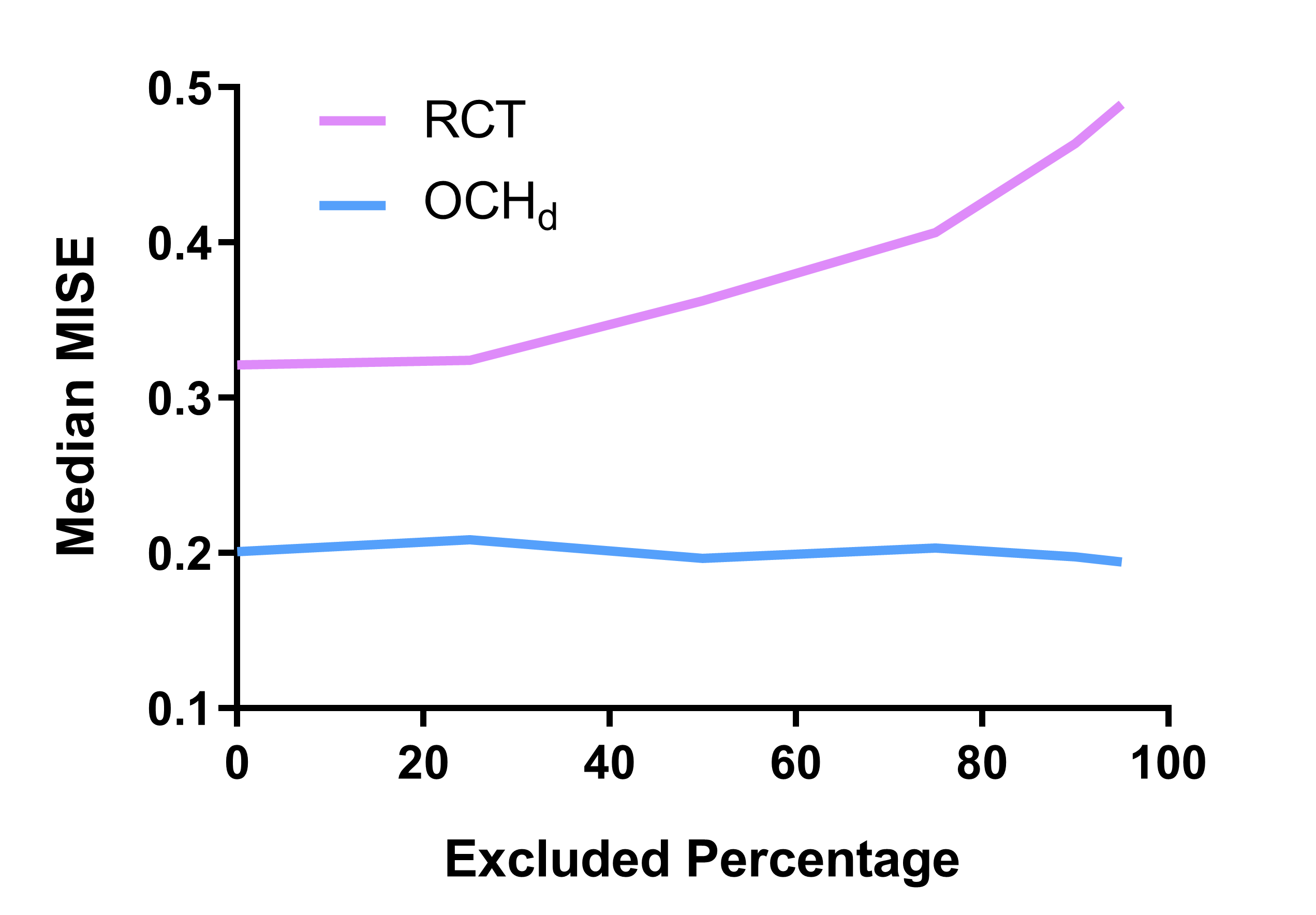}
    \caption{} \label{fig:OCHdv}
\end{subfigure}
\caption{Stability results when Assumptions \ref{assump:main_e} and \ref{assump:main_p} are violated. Results are qualitatively similar to those in Figure \ref{fig:OCH}.} \label{fig:viol:stab}
\end{figure}

We repeat the simulation experiments but consider violations of Assumptions \ref{assump:main_e} and \ref{assump:main_p}. Let $\delta_T = E(Y_1(T)|T,\bm{X}) - E(Y_0|\bm{X})$. We sample the RCT data with probability 0.5 from:
\begin{equation*}
\begin{aligned}
    &Y_1(T) \sim \mu_T \mathcal{N}(f_{1T}(Z) + \delta_T,0.1) + (1-\mu_T)\mathcal{N}(f_{1{T}}(Z),0.1),
\end{aligned}
\end{equation*}
and with the other probability 0.5 from:
\begin{equation*}
\begin{aligned}
    &Y_1(T) \sim \mu_T \mathcal{N}_2(Z) + (1-\mu_T)\mathcal{N}^\prime_2(Z),
\end{aligned}
\end{equation*}
where $\mathcal{N}^\prime_2(Z) = \frac{1}{2}\mathcal{N}(f_{01}(Z),0.1) + \frac{1}{2} \mathcal{N}(f_{00}(Z),0.1) -\delta_T$. Pictorially, these situations correspond to sampling from outside the assumed convex hull:
\begin{center}
\begin{tikzpicture}[mydrawstyle/.style={draw=black, very thick}, x=1mm, y=1mm, z=1mm]
  \draw[mydrawstyle, ->](-2,30)--(66,30);
  
  \draw[mydrawstyle](20,28)--(20,32) node[below=10]{};
  \draw[mydrawstyle](40,28)--(40,32) node[below=10]{};
  
  \fill[fill=blue, opacity=0.2](0,29)--(0,31)--(20,31)--(20,29)--(0,29);
  \draw[mydrawstyle, draw=blue](1,29)--(0,29)--(0,31)--(1,31);
   \draw[mydrawstyle, draw=blue](19,29)--(20,29)--(20,31)--(19,31);

     \fill[fill=blue, opacity=0.2](40,29)--(40,31)--(60,31)--(60,29)--(40,29);
  \draw[mydrawstyle, draw=blue](41,29)--(40,29)--(40,31)--(41,31);
   \draw[mydrawstyle, draw=blue](59,29)--(60,29)--(60,31)--(59,31);
  
  \node[] at (20, 18.5)  (zero)     {\begin{tabular}{c}\smaller{$\mathbb{E}^o(Y_0|\bm{X})$}\end{tabular}};
  \node[] at (40, 18.5)  (obs)     {\begin{tabular}{c}\smaller{$\mathbb{E}^o(Y_1|\bm{X},T)$}\end{tabular}};

    \draw [decorate,
    decoration = {brace,amplitude=10pt}] (0,33)--(20,33) node at (10,39.5) {\begin{tabular}{c} \smaller{$-\delta_T$} \end{tabular}};

        \draw [decorate,
    decoration = {brace,amplitude=10pt}] (40,33)--(60,33) node at (50,39.5) {\begin{tabular}{c} \smaller{$+\delta_T$} \end{tabular}};
  
  \path [->] (zero) edge node[sloped,bend left] {}(20,26.75);
  \path [->] (obs) edge node[sloped,bend left] {}(40,26.75);
\end{tikzpicture}
\end{center}
\noindent We summarize accuracy results in Tables \ref{table:add:CATE} and \ref{table:add:CDTE}, and stability results in Figure \ref{fig:viol:stab}.

\begin{table}[h]
\centering
\begin{subtable}{0.98\textwidth}  
\centering
\captionsetup{justification=centering,margin=2cm}
\begin{tabular}{ccccccc}
\hhline{=======}
      & 0\%             & 25              & 50              & 75              & 90              & 95              \\ \hline
\rowcolor{Gray} UNC$_1$   & \textbf{0.1580} & \textbf{0.1698}          & \textbf{0.2162}          & \textbf{0.2314}          & \textbf{0.2729}      & 0.3226\\
\rowcolor{Gray} UNC$_2$   & \textbf{0.1471} & \textbf{0.1667}          & \textbf{0.2259}          & \textbf{0.2598}          & 0.3248          & 0.3540\\
\rowcolor{Gray} OCH$_2$   & 0.1892 & \textbf{0.1878} & \textbf{0.2106} & \textbf{0.2418} & \textbf{0.2730} & \textbf{0.2818} \\
\rowcolor{Gray} OCH$_1$  & 0.2017          & 0.2000          & \textbf{0.2008}         & \textbf{0.2280}         & \textbf{0.2377}         & \textbf{0.2553}          \\
SDD   & 0.2643          & 0.2659          & 0.3061         & 0.3222          & 0.3693          & 0.3898          \\
2Step & 0.3407          & 0.3417          & 0.3677         & 0.3494          & 0.3705          & 0.3820          \\
OBS   & 0.3707          & 0.3882          & 0.3793          & 0.3659          & 0.4010          & 0.3733         \\
RCT   & 0.2928          & 0.3309          & 0.4457          & 0.6141          & 0.7680          & 0.7425          \\
CDD   & 0.8735          & 0.8356          & 0.8895          & 0.8491          & 0.8895         & 0.8983          \\ 
OLT   & 0.5674          & 0.6011          & 0.7160          & 0.8168         & 0.9671          & 1.0516          \\
\hhline{=======}
\end{tabular}
\caption{} \label{exp:CATE:propv}
\end{subtable}

\vspace{3mm}\begin{subtable}{0.98\textwidth}  
\centering
\captionsetup{justification=centering,margin=2cm}
\begin{tabular}{ccccc}
\hhline{=====}
      & 1               & 3               & 6               & 10              \\  \hline
\rowcolor{Gray} UNC$_2$   & 0.1329         & \textbf{0.1279}         & \textbf{0.2927} & \textbf{0.4822}\\
\rowcolor{Gray} UNC$_1$   & \textbf{0.0999}          & \textbf{0.1265}          & \textbf{0.3036} & \textbf{0.4857}\\
\rowcolor{Gray} OCH$_2$   & \textbf{0.0779} & \textbf{0.1453} & \textbf{0.3313} & \textbf{0.5392} \\
\rowcolor{Gray} OCH$_1$   & \textbf{0.0763}         & 0.1488          & \textbf{0.3375}          & \textbf{0.5361}          \\
SDD   & 0.1106          & 0.2017          & 0.4388          & 0.6770          \\
2Step & 0.0695          & 0.2288          & 0.5481          & 0.9180          \\
RCT   & 0.1935          & 0.3347          & 0.5576          & 0.8427          \\
OBS   & 0.0800          & 0.2361          & 0.6107          & 1.0483         \\
OLT   & 0.5561          & 0.6117          & 0.9586          & 1.2268          \\
CDD   & 0.3054          & 0.6240          & 1.1115          & 1.7799          \\  
\hhline{=====}
\end{tabular}
\caption{}  \label{exp:CATE:varsv}
\end{subtable}
\caption{Accuracy results for the CATE when Assumption \ref{assump:main_e} is violated. Sub-figures (a) and (b) correspond to different percentages of excluded subjects and numbers of variables in $\bm{X}$, respectively. The two algorithms OCH$_2$ and OCH$_1$ are only outperformed by their unconstrained variants when Assumption \ref{assump:main_e} fails to hold.} \label{table:add:CATE}
\end{table}

\begin{table}[h]
\begin{subtable}{0.98\textwidth}  
\centering
\captionsetup{justification=centering,margin=2cm}
\begin{tabular}{ccccccc}
\hhline{=======}
      & 0\%             & 25              & 50              & 75              & 90              & 95              \\ \hline
\rowcolor{Gray} OCH$_d$   & \textbf{0.2008} & \textbf{0.2084} & \textbf{0.1964} & \textbf{0.2030} & \textbf{0.1973} & \textbf{0.1940} \\
\rowcolor{Gray} UNC$_d$   & \textbf{0.2047} &  \textbf{0.2134}          & \textbf{0.2063}          & \textbf{0.2106}          & \textbf{0.2056}          &  \textbf{0.2052}\\
OBS   & 0.2341          & 0.2405          & 0.2460          & 0.2379          & 0.2366         & 0.2372          \\
RCT   & 0.3210          & 0.3242          & 0.3625          & 0.4064          & 0.4636          & 0.4890          \\
\hhline{=======}
\end{tabular}
\caption{} \label{table:CDTE:propv}
\end{subtable}

\vspace{3mm} \begin{subtable}{0.98\textwidth}  
\centering
\captionsetup{justification=centering,margin=2cm}
\begin{tabular}{ccccc}
\hhline{=====}
      & 1               & 3               & 6               & 10              \\  \hline
\rowcolor{Gray} OCH$_d$   & \textbf{0.0687} & \textbf{0.1620} & \textbf{0.2361} & \textbf{0.3016} \\
\rowcolor{Gray} UNC$_d$   & 0.0864          & \textbf{0.1649}          & \textbf{0.2391} & \textbf{0.3068}\\
OBS   & 0.0992          & 0.1993          & 0.2714          & \textbf{0.3104}          \\
RCT   & 0.3255          & 0.3755          & 0.3869          & 0.4118         \\
\hhline{=====}
\end{tabular}
\caption{} \label{table:CDTE:varsv}
\end{subtable}
\caption{Accuracy results for the CDTE when Assumption \ref{assump:main_p} does not hold. OCH$_d$ outperforms RCT and observational data only for all (a) percentages of excluded subjects, and (b) numbers of variables in $\bm{X}$.}  \label{table:add:CDTE}
\end{table}

\newpage
\subsection{Additional Real Data Results} \label{app:real:add}

Many randomized trials exclude children, but children get sick too. We therefore evaluated the algorithms on how well they generalize to children when trained on an RCT only recruiting adults and a confounded observational dataset.

\begin{figure}[h]
\begin{subfigure}[]{0.5\textwidth}
    \centering
    \captionsetup{justification=centering,margin=2cm}
    \includegraphics[scale=0.65]{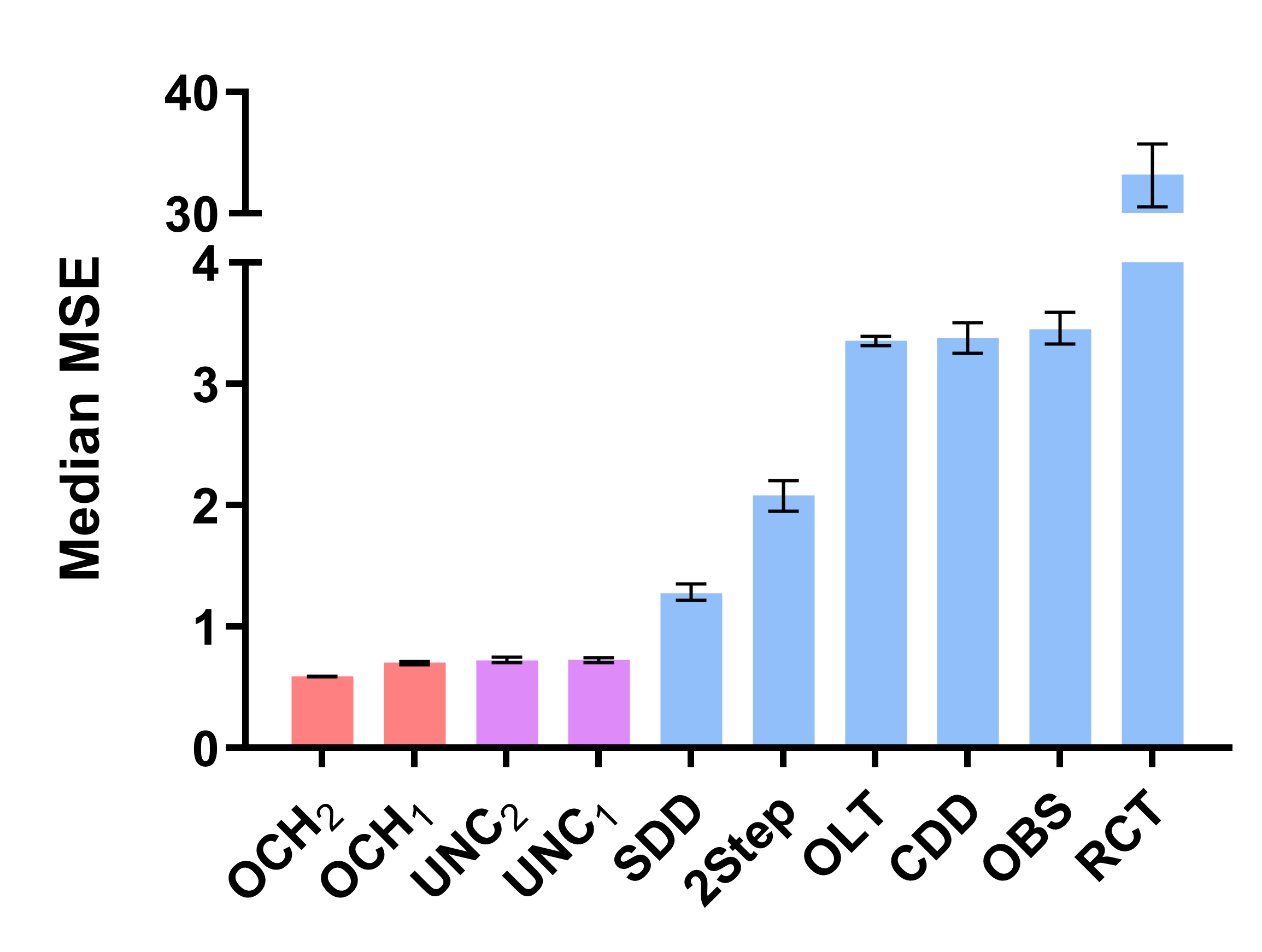}
    \caption{} \label{fig:TEOSS}
\end{subfigure}
\begin{subfigure}[]{0.5\textwidth}
    \centering
    \captionsetup{justification=centering,margin=2cm}
    \includegraphics[scale=0.65]{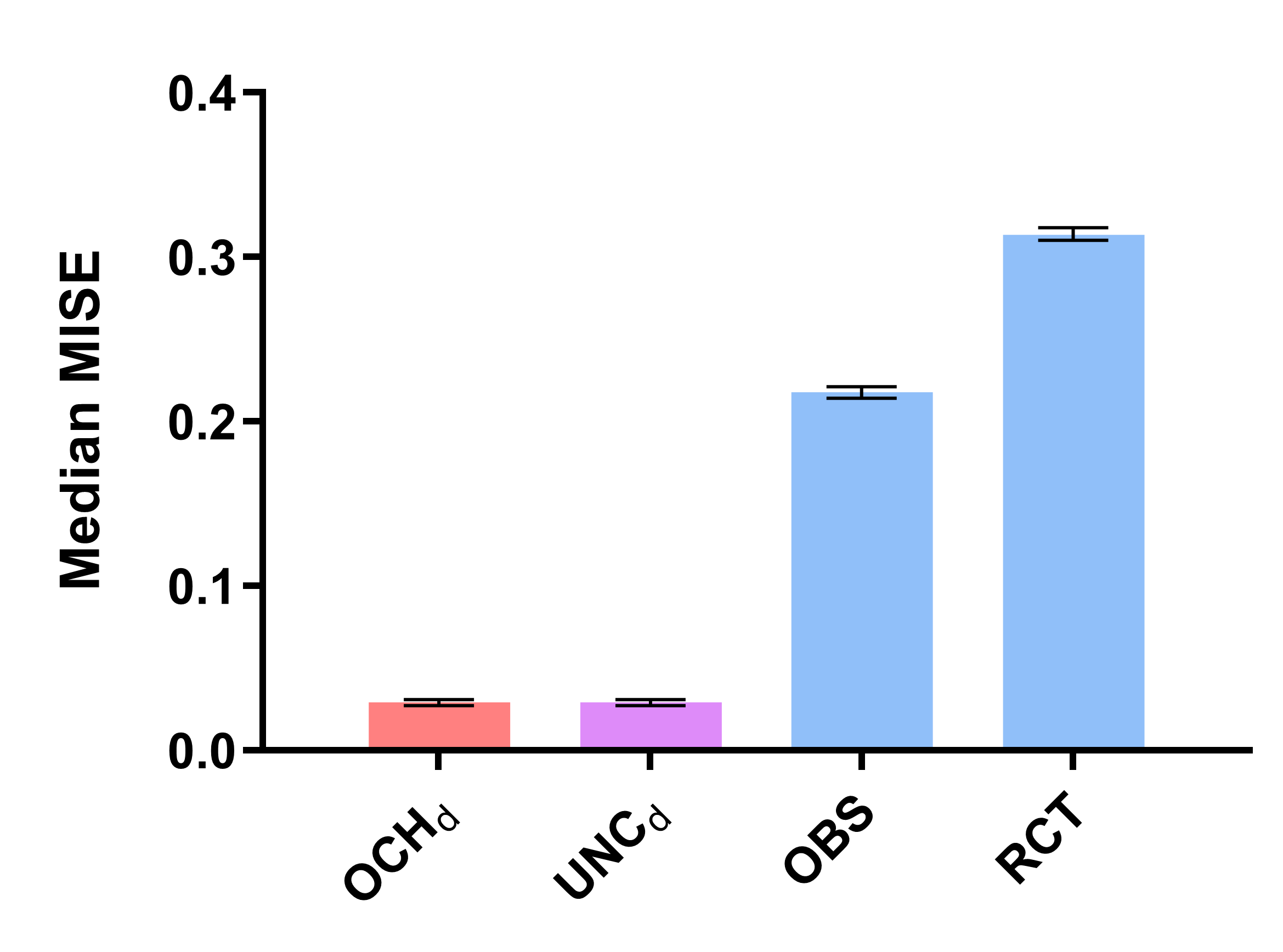}
    \caption{} \label{fig:TEOSS:dens}
\end{subfigure}
\caption{The (a) median MSE values and (b) median MISE values for CATIE and TEOSS. The OCH algorithms again achieve the best performance.} \label{fig:real:app}
\end{figure}

We in particular obtained data from two clinical trials investigating the effects of anti-psychotics on schizophrenia spectrum disorders. The CATIE trial recruited 530 adults who were at least 18 years old, while the TEOSS trial recruited 62 children up to 19 years old \citep{Mcevoy05,Sikich08,Stroup09}. 

Clinicians prefer olanzapine ($T=1$) over risperidone ($T=0$) for excited patients -- defined as hyperactivity, heightened responsivity, hyper vigilance or excessive mood lability -- because olanzapine is more sedating. We therefore set the excitement subscore of the PANSS scale, a quantitative measure of schizophrenia symptoms, at week 4 as the outcome. We then predicted differences in excitement using age and the PANSS hostility sub-score as predictors in $\bm{X}$ because adults who are hostile are more dangerous than weaker children. The comprehensive RCT corresponds to the combined CATIE and TEOSS dataset consisting of $530+62 = 592$ patients.

We generated the \textcolor{blue}{observational data} by first combining CATIE and TEOSS. We then excluded patients assigned to $T=1$ who were not excited (excitement subscore less than or equal to two -- the formal cutoff for questionable pathology), and patients assigned to $T=0$ who were excited (subscore greater than two). This mimics real world prescribing patterns where physicians prescribe olanzapine to excited patients.

We used the 530 patients in the CATIE trial as the \textcolor{blue}{exclusive RCT data}. Since we include age as a predictor, the goal is to generalize to children. We therefore tested the algorithms on their ability to accurately predict the CATE or CDTE on children using this adult RCT and the confounded observational dataset.

We report the results over 2000 bootstrapped draws in Figures \ref{fig:TEOSS} and \ref{fig:TEOSS:dens}. OCH$_2$ and OCH$_1$ again outperformed their predecessors and their unconstrained variants (Figure \ref{fig:TEOSS}). Similar results held with OCH$_d$, although OCH$_d$ did not outperform UNC$_d$ in this case similar to the synthetic data results (Figure \ref{fig:TEOSS:dens}). The performance improvements were much larger in this dataset compared to STAR*D. Finally notice that the RCT only algorithm performs terribly (median MSE 33.2 and median MISE 0.31) because non-linear regressors or conditional density estimators cannot consistently extrapolate well from adults to children, or to unseen regions on the support $S_O \setminus S_R$.

\end{document}